\documentclass[]{trbunofficial}
\usepackage{graphicx}

\usepackage{comment}
\usepackage{amsmath}

\usepackage{color}
\usepackage{subcaption}
\usepackage{booktabs}
\usepackage{multirow}
\usepackage{array}
\usepackage{wrapfig}

\definecolor{corner}{RGB}{33,102,172}
\definecolor{sidewalk}{RGB}{239,138,98}
\definecolor{crossing}{RGB}{178,24,43}
\definecolor{classtext}{RGB}{253,219,199}

\definecolor{Red}{RGB}{225, 25, 5}
\definecolor{Green}{RGB}{51, 160, 44}
\definecolor{Blue}{RGB}{0, 100 ,200}
\definecolor{Yellow}{RGB}{255, 255 ,0}

\definecolor{BrightRed}{RGB}{225, 0, 0}
\definecolor{LightYellow}{RGB}{254, 254 ,139}

\usepackage[hidelinks]{hyperref}

\AuthorHeaders{Zhang et al.}
\title{Reliable, Routable, and Reproducible: Collection of Pedestrian Pathways at Statewide Scale}

\author{%
  \textbf{Yuxiang Zhang, Bill Howe, Anat Caspi}\\
  University of Washington\\
\{yz325,billhowe,caspian\}@uw.edu
}

\begin{document}
\maketitle

\section{Abstract}

While advances in mobility technology including autonomous vehicles and multi-modal navigation systems can improve mobility equity for people with disabilities, these technologies depend crucially on accurate, standardized, and complete pedestrian path networks. Ad hoc collection efforts lead to a data record that is sparse, unreliable, and non-interoperable. 

This paper presents a sociotechnical methodology to collect, manage, serve, and maintain pedestrian path data at a statewide scale. Combining the automation afforded by computer-vision approaches applied to aerial imagery and existing road network data with the quality control afforded by interactive tools, we aim to produce routable pedestrian pathways for the entire State of Washington within approximately two years. We extract paths, crossings, and curb ramps at scale from aerial imagery, integrating multi-input segmentation methods with road topology data to ensure connected, routable networks. We then organize the predictions into project regions selected for their value to the public interest, where each project region is divided into intersection-scale tasks. These tasks are assigned and tracked through an interactive tool that manages concurrency, progress, feedback, and data management. 

We demonstrate that our automated systems outperform state-of-the-art methods in producing routable pathway networks, which then significantly reduces the time required for human vetting.  Our results demonstrate the feasibility of yielding accurate, robust pedestrian pathway networks at the scale of an entire state. 

This paper intends to inform procedures for national-scale ADA compliance by providing pedestrian equity, safety, and accessibility, and improving urban environments for all users.

\hfill\break%
\noindent\textit{Keywords}:  Pedestrian, Sidewalk, Walkability, Accessibility, Machine Learning, Automated System

\section{Introduction}
\label{sec:intro}


The importance of comprehensive and accurate pedestrian and bicycle data is increasingly evident within the transportation planning and technology sectors. As cities and communities strive to enhance mobility, accessibility, safety, and sustainability, the need for standardized, reliable, national-scale data pedestrian and bike data has emerged as a critical priority. 


Pedestrian networks serve as vital links that connect pedestrians to various modes of transportation, ensuring seamless mobility. However, the current state of pedestrian path information is often sparse and unreliable due to ad hoc, site-specific data collection efforts, typically conducted without adherence to standardized protocols, and with minimal consideration for routability and quality to ensure downstream utility. 

To address these issues, we present a sociotechnical protocol for scalable and accurate data collection that integrates automated predictions from aerial imagery with interactive expert vetting and community engagement inclusive of the people most affected by data quality. Specifically, the automated predictions are produced by the Prophet system, which consists of three components: First, a state-of-the-art vision model was adapted to accept both aerial imagery and rasterized road networks as input and produce aerial images segmented into sidewalks, crossings, and corner bulbs as output.  Second, the system uses publicly available road network data to heuristically hypothesize the location of the predicted elements. Third, Prophet uses labeled, segmented images to optimize the locations of the hypothesized elements. The result is a predicted network that tends to be more connected, thereby significantly reducing the time required for manual vetting and improving utility in downstream routing applications.

Our manual vetting protocol, which we call Skeptic, consists of three stages: First, we identify a project area and divide it into intersection-scale tasks that can be vetted in just a few minutes (in some cases, seconds.) Second, trained assessors use an online interactive tool to lock an intersection task, repair the predicted elements so that they better match the underlying aerial imagery, then commit their changes to the system. Third, community assessment teams physically review the built environment in selected areas for further validation and to augment the data with additional features relevant to their local needs (e.g., the existence of accessible pedestrian signaling). This process affords multiple redundancies for self-validation both between stages (model, expert, and community) and within stages (e.g., assigning the same task to multiple experts, or adopting overlapping project areas). Going forward, we are implementing feedback mechanisms such that the model can be continuously fine-tuned using expert vetting, and the experts can learn from community-specific needs.  Our goal is to refine this overall process to ensure efficient, reliable data collection at national scales.   

In this paper, we consider the problem, background, and related work, then describe an end-to-end system consisting of Prophet and Skeptic. We then present experiments showing that Prophet outperforms imagery-only systems in producing routable networks, and that Skeptic further improves connectivity with modest manual effort. 
Our results indicate that this process yields highly accurate and robust pedestrian pathway networks across multiple states, showcasing the potential for nationwide application. The establishment of a national specification for pedestrian and bike data will not only enhance the efficiency of transportation systems but also promote safer and more accessible urban environments for all users.



\section{Problem, Background, and Related Work}
\label{sec:related}

Even after 35 years, ADA compliance in the public right of way remains remarkably limited. The ADA requires identifying and removing mobility barriers on the ground, which in turn requires comprehensive, accurate, low-cost data collection to overcome the institutional, cultural, and funding obstacles impeding compliance~\cite{wagner2024navigating}. 

These data are exceptionally challenging to collect relative to, for example, road networks. A pedestrian map is not simply a set of lines: it must also represent connectivity, transitions, and additional attributes to afford navigation, particularly for those with disabilities. These elements vary significantly by jurisdiction, zoning, and even year built; the heterogeneity makes the process exceptionally error-prone even for experts~\cite{bolten2021towards, bolten2022towards}.
Proposed solutions involve some combination of direct inspection of the built environment, computational inference of remotely sensed data, or interactive review of existing sources. Direct inspection can provide highly detailed and accurate data with sufficient training but does not scale due to prohibitive costs.  
Computational inference scales broadly but struggles with unusual situations and occlusions, leading to biased outputs \cite{rhoads2023sidewalk}.  Interactive repair combines automated and manual approaches, allowing for modest-scale data collection and quality control, but remains sensitive to the training and experience of the operators.  We argue that all three approaches, organized interdependently, offer the only viable solution.
Prior work has recognized the need for combining multiple sources of information. 
Wu et al.  \cite{wu2019road} used OpenStreetMap (OSM) centerlines as labeled data and extracted roads from very high-resolution (VHR) satellite images. Sun et al. \cite{sun2019leveraging} added crowd-sourced global positioning system (GPS) data to satellite images to extract roads with CNN-based semantic segmentation. Zhou et al. \cite{zhou2021funet} fused remote sensing images and GPS for road detection and extraction. Additional recent learning-based studies include Lu et al.  \cite{lu2021gamsnet} proposing a multi-scale residual neural architecture for road detection, Pan et al. \cite{pan2021generic} proposing a fully convolutional neural network using VHR remote sensing, Mattyus et al. \cite{mattyus2017deeproadmapper} estimated road topology from aerial images, Mi et al. \cite{mi2021hdmapgen} generated road lane graphs from LiDAR data with a hierarchical graph generation model. However, all of these methods focus on automobile road extraction and ignore pedestrian paths entirely.  



Crowdsourcing offers another approach to the scale problem. WheelMap \cite{mobasheri2017wheelmap} collects subjective wheelchair accessibility information about businesses and public venues and submits it to OpenStreetMap.
Project Sidewalk \cite{saha2019project} crowdsources data on accessibility issues within the pedestrian infrastructure by labeling point locations based on Google StreetView imagery.
These efforts tend to lack standards for data completeness, quality, interpretation, and open availability, limiting their downstream reliability.

Recent advancements in remote sensing have led to methods to automate pedestrian path data collection. Ahmetovic et al. \cite{ahmetovic2015zebra} detect zebra crossings using satellite imagery validated with street-level images. Ghilardi et al. \cite{ghilardi2016crosswalk} classified and located crosswalks using an SVM classifier over data extracted from road maps, and Ning et al. \cite{ning2022sidewalk} extracted sidewalks from aerial images with a neural network, restoring occluded segments from street view images. While these studies have improved pedestrian environment mapping, they do not generate a comprehensive, connected, and routable pedestrian pathway network graph necessary for city planning and navigation. Tile2Net \cite{hosseini2023mapping} infers sidewalk, crosswalk, and footpath centerlines at scale from satellite imagery, but ignores connectivity entirely. Li et al. \cite{li2018semi} developed a semi-automated method to generate sidewalk networks using parcel-level data and roadway centerlines, but accuracy was very limited without human editing; we incorporate a similar approach into the Prophet system to extract hypothesized sidewalks that are improved using computer vision approaches, followed by a principled manual vetting process.

Other studies rely on collecting and review of on-the-ground data. PathVu \cite{pathVu} collects sidewalk path information but does not maintain path connectivity, limiting utility for routing applications.
Zhang et al. \cite{zhang2021collecting} automated the collection of street-view images with auxiliary data to map sidewalk connectivity and infrastructure, and
Hou et al. \cite{hou2020network} extracted sidewalk paths using LiDAR data and point cloud segmentation. These methods all involve the deployment of (sometimes proprietary) physical sensing platforms, limiting their scalability. 

To truly advance accessibility and ensure compliance at scale, we must also tackle institutional, organizational, and cultural barriers that hinder progress~\cite{wagner2024navigating, eisenberg2024planning}. Together, Prophet and Skeptic not only produce high-quality predictions of pedestrian pathways and organize expert attention to vet these predictions, but also afford community engagement for ground validation by those most affected by the data. This work explores the multifaceted sociotechnical approaches needed to keep stakeholders engaged, meeting them where they are, as the overall systems become increasingly automated to achieve comprehensive, reliable, and scalable accessibility improvements.

\begin{figure}[t!]
\centering
\includegraphics[width=\columnwidth]{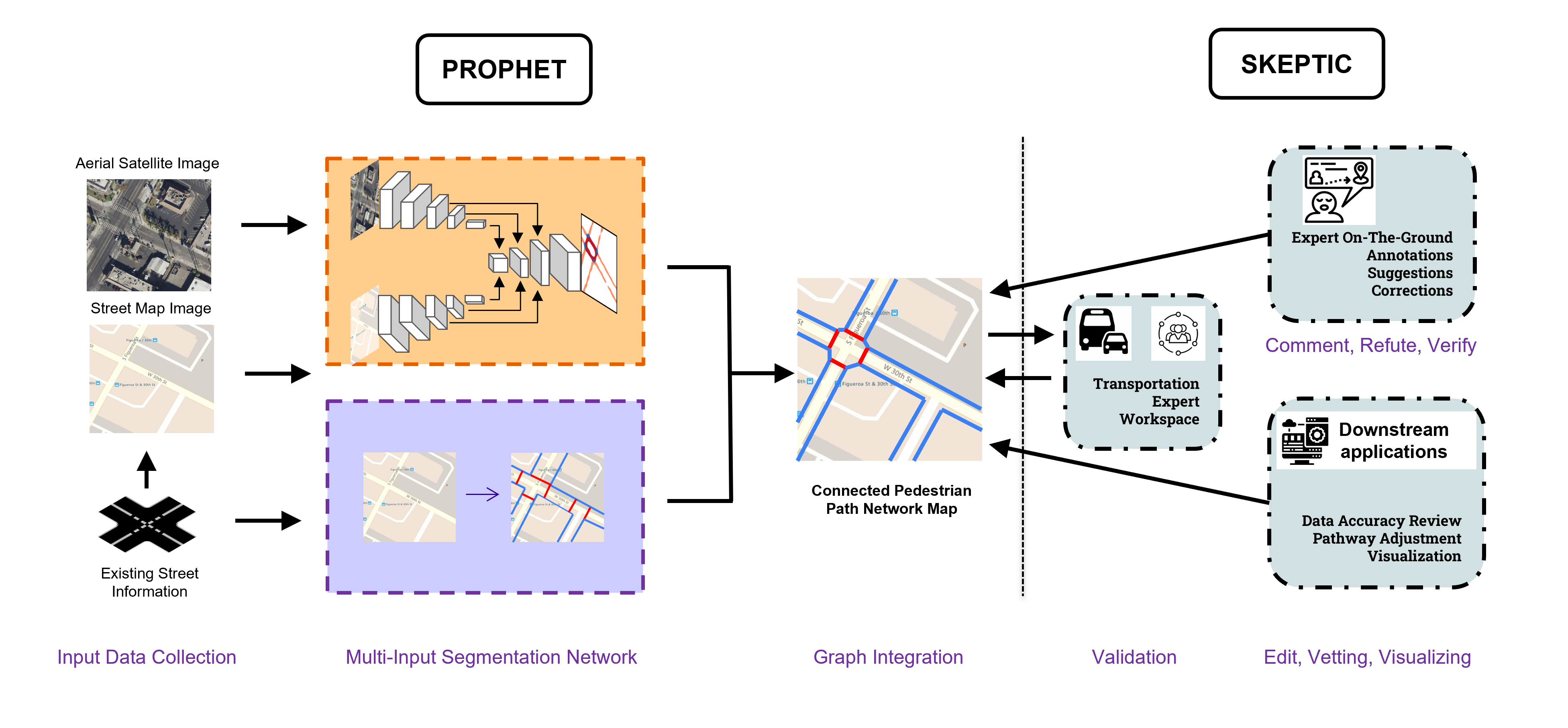}
\caption{Overview of the end-to-end process for inferring pedestrian path network graph: Section \ref{sec:Prophet} describes Prophet, the multi-input generating a pedestrian path network from incomplete information. Section \ref{sec:cnn} details the segmentation network using aerial and road map images to predict class labels of pixel locations in the pedestrian map. Section \ref{sec:opt} explains how \textit{Pedestrianfer} and segmentation network information are combined for accurate graph inference. Section \ref{sec:edits} describes the Skeptic protocol for manual vetting and community engagement.}
\label{fig:flow_chart}
\end{figure}

\section{Prophet: Network-Centered Inference of Pedestrian Pathway Networks}
\label{sec:Prophet}
In this section, we introduce our fully automated method for pedestrian path network graph inference called Prophet. As shown in Figure \ref{fig:flow_chart}, the process consists of three main steps. First, we developed a module called \textit{Pedestrianfer} to create a hypothesized graph with existing street network data. Second, we use a multi-input segmentation network to generate pixel-wise prediction masks for the important classes in the pedestrian environment. Lastly, we use the information from the segmentation network to optimize the hypothesized graph into an accurate connected pedestrian path network graph.

\subsection{Pedestrianfer}
\label{sec:pedestrianfer}
\begin{wrapfigure}{r}{0.42\textwidth}
\centering
\includegraphics[width=0.39\columnwidth]{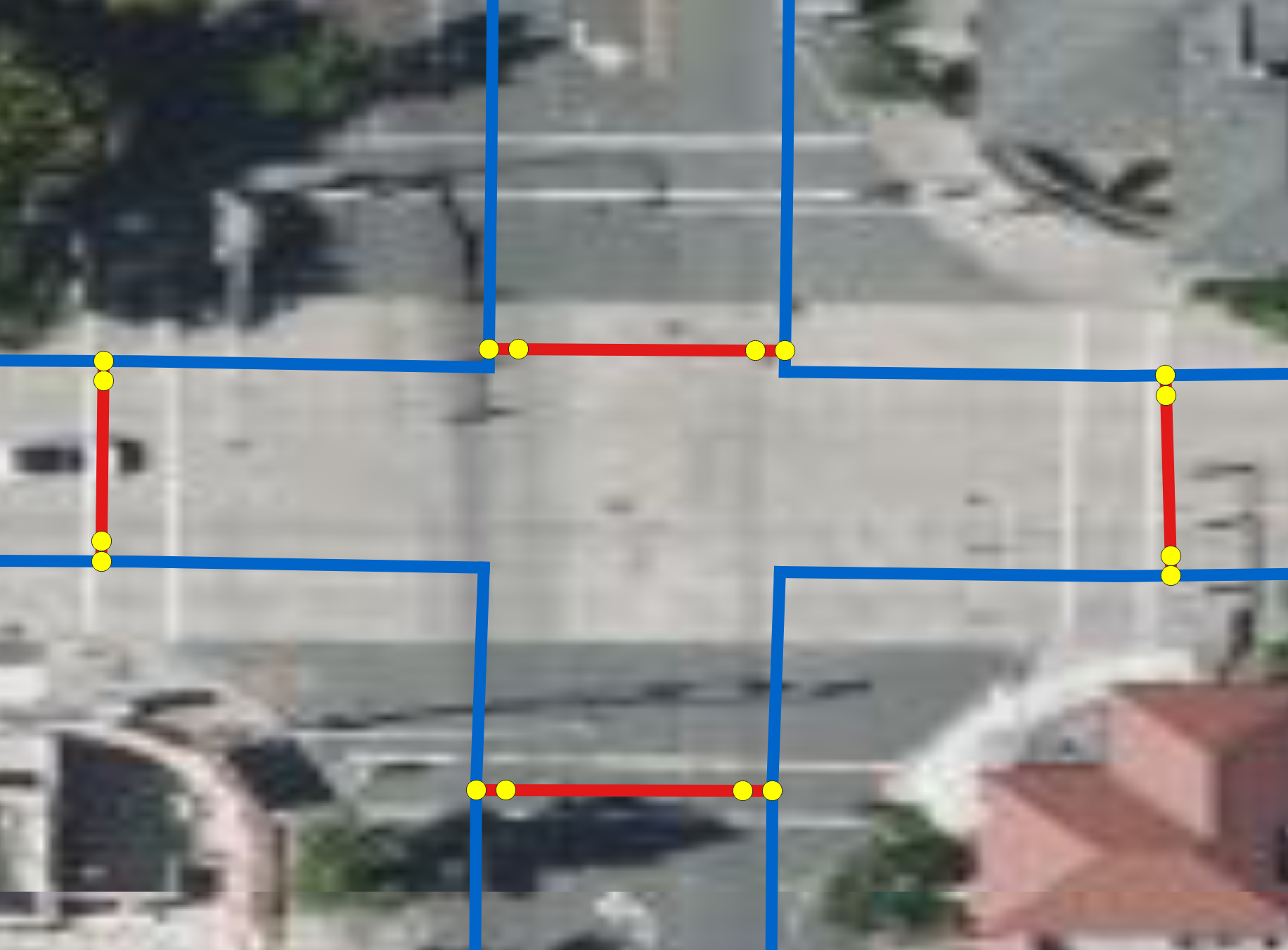}
\caption{Hypothesized graph generated with \textit{Pedestrianfer}. \colorbox{Yellow}{Yellow} dots represent the hypothesized nodes for the curb modes and the nodes on the sidewalk. \colorbox{Blue}{Blue} lines represent the hypothesized sidewalks. \colorbox{Red}{Red} lines represent the hypothesized crossings. Links are part of the crossings between the curb nodes and the nodes on the sidewalk.}
\label{fig:hypo}
\end{wrapfigure}

To infer a hypothesized pedestrian path network from incomplete information, we developed \textit{Pedestrianfer} (Figure \ref{fig:hypo}) which (1) infers an (optimistic) sidewalk network from existing street networks, (2) infers street crossing locations from a street network and a sidewalk network, and (3) infers curb transitions between crossings and sidewalks. \textit{Pedestrainfer} is similar to previous work \cite{li2018semi} but with the following key differences: (1) the only required input is a street network
(2) full intersection-to-intersection sidewalk paths are estimated rather than assuming 50-meter segments, and (3) street crossing pathways are generated via a cost function that does not require manual intervention. Additionally, Pedestrianfer generates pathways according to the OpenSidewalks Schema specification \cite{opensidewalksSchema}.

\textit{Pedestrianfer} first infers sidewalk networks from a street network under two alternative regimes, 
depending on whether the data source 
includes metadata on the presence (and optionally, offset distance) of sidewalks. If metadata is present, sidewalks are placed only where indicated. If not, \textit{Pedestrianfer} retrieves a street network from OSM and hypothesizes full sidewalks, i.e. the scenario in which all streets have a connected sidewalk on either side. 

In either case, \textit{Pedestrianfer} involves 4 steps: (1) creating a directed graph representation of the street network, (2) generating all right-hand-turn paths that start and stop at the same node (closed paths), (3) drawing sidewalks via a line offset algorithm, and (4) trimming or joining them based on the path context, e.g. trimming overlapping sidewalk lines along the path.

Next, \textit{Pedestrianfer} uses the street and sidewalk networks 
to infer crossings. In this step, \textit{Pedestrianfer} iterates over each  intersection node in the street network and generates (1) all street lines associated with the intersection, and (2) a set of candidate sidewalks to connect with a crossing on each side of each street. \textit{Pedestrianfer} then generates candidate crossings as lines drawn from a sidewalk on the left side of a street to a sidewalk on the right side of the street. Multiple candidates are generated for each street of an intersection by selecting a series of points along that street and generating a crossing that connects the closest corresponding left and right sidewalks. Metrics of the candidates are generated, including the distance of the street point to the intersection, the crossing line length, and the angle between the crossing line and the street it crosses. A cost function is then used to heuristically select the crossing that minimizes a linear combination of the distance to the street intersection, crossing line length, and non-orthogonal crossing angles. Therefore, \textit{Pedestrianfer} estimates street crossing locations that are near intersections, which align with common policies and pedestrian safety measures. In the case that crossing locations with ground markings are already known and present in an associated dataset (as point data), \textit{Pedestrianfer} will generate a crossing near that point by projecting to the nearest known sidewalk candidates on each side.

Lastly, \textit{Pedestrianfer} splits crossing lines into three segments: (1) the originating sidewalk surface, (2) the street surface, and (3) the destination sidewalk surface. These correspond to typical surfaces a pedestrian would travel in an urban area. Sidewalk-street transitions, which often have vertical displacements or are where curb-cut ramps meet the street, can be interpreted as potential curb interfaces. Estimates produced by \textit{Pedestrianfer} 
for sidewalk-curb-street locations
can then be improved with manual mapping 
or learning-based methods as discussed below.

\subsection{Segmentation Network}
\label{sec:cnn}
In order to verify, correct, and refine the \textit{Pedestrianfer} hypothesized path network graph, we use inference from a segmentation network. The segmentation network is trained on the PathwayBench dataset~\cite{zhang2024pathwaybench}, which contains aerial satellite imagery and co-registered rasterized street map as inputs, and provides annotations to important classes in the pedestrian environment including, \textit{sidewalk}, \textit{crossing}, and \textit{corner bulb}. The segmentation network 
uses information from both the aerial satellite image and the street map tiles. As shown in Figure \ref{fig:flow_chart}, the segmentation network has two identical branches: the aerial satellite images are used as input in one branch and street map tiles are used in the other. Each branch has its encoder-decoder structure. To fuse the information from both branches, we concatenate feature maps from both branches at different layers hierarchically during the up-sample process. We discuss the use of different models as the segmentation backbone models and their performance in the \nameref{sec:exp} section.

\subsection{Optimization for Full Pedestrian Path Networks}
\label{sec:opt}

\textit{Pedestrianfer} generates a hypothesized pathway graph that outlines the potential location and connectivity of sidewalks and crossings. To obtain a more accurate pedestrian pathway graph, we use information from the segmentation network to refine the hypothesized pathway graph. Our strategy is to first find the optimized geolocation of the nodes in the graph, then connect the nodes with edges to complete the graph.  

\begin{wrapfigure}{r}{0.65\textwidth}
    \centering
        \begin{subfigure}[]{0.25\columnwidth}
        \centering
        \caption{}
        \includegraphics[width=\columnwidth]{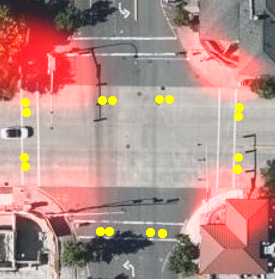}
        \label{fig:node_hypo}
    \end{subfigure}
    \begin{subfigure}[]{0.25\columnwidth}
        \centering
        \caption{}
        \includegraphics[width=\columnwidth]{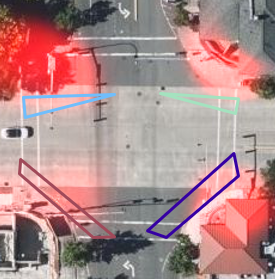}
        \label{fig:node_before}
    \end{subfigure}
    \begin{subfigure}[]{0.25\columnwidth}
        \centering
        \caption{}
        \includegraphics[width=\columnwidth]{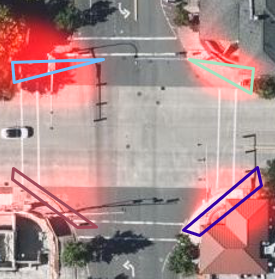}
        \label{fig:node_after}
    \end{subfigure}
    \begin{subfigure}[]{0.25\columnwidth}
        \centering
        \caption{}
    \includegraphics[width=\columnwidth,height=118px]{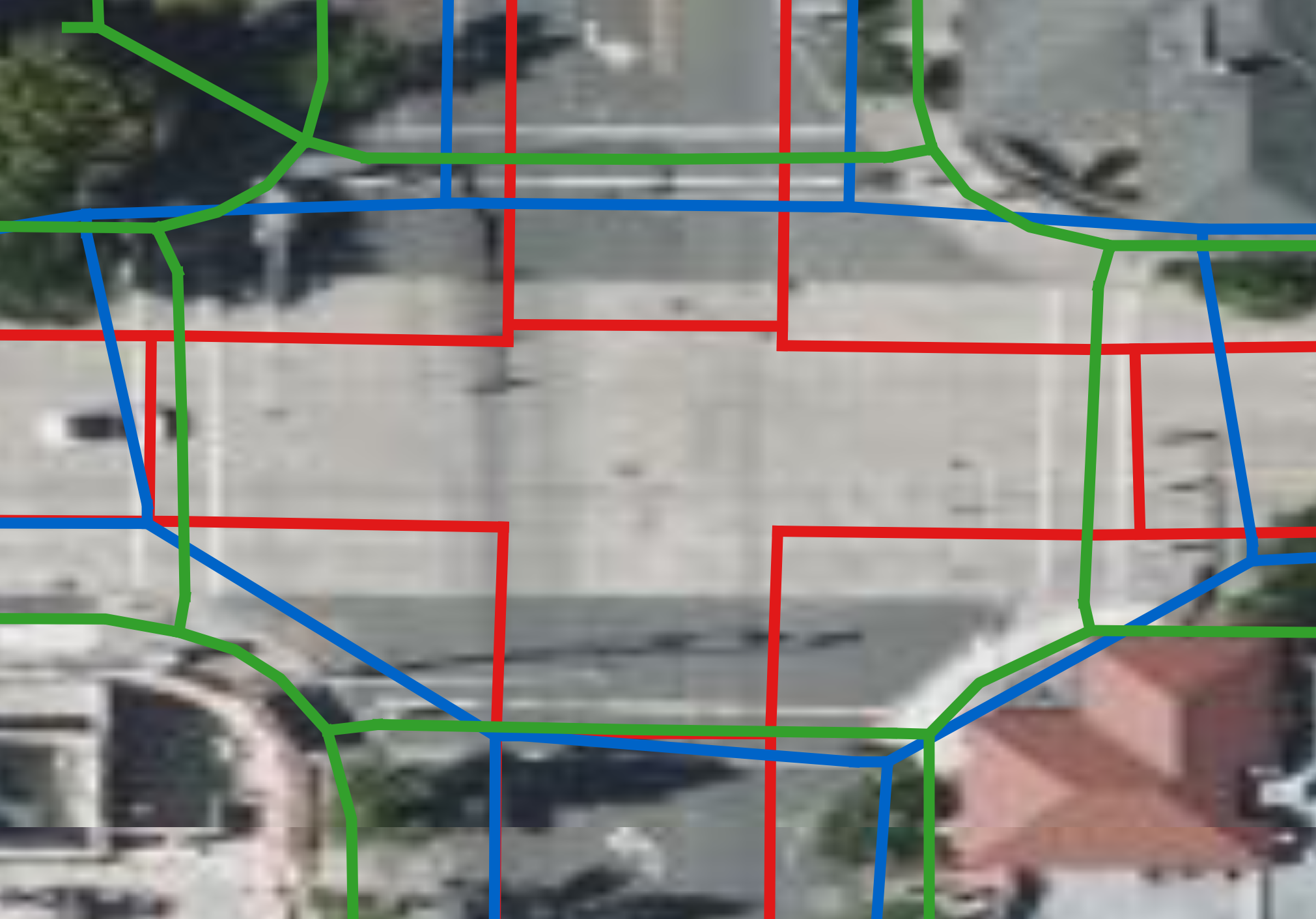}
        \label{fig:node_graph}
    \end{subfigure}

    \caption[Illustration of node location optimization in Prophet]{Illustration of node location optimization, the probability of each pixel being the \textit{corner bulb} is shown as a heat map in red in (a) - (c). (a) \textit{Pedestrianfer} hypothesized nodes (b) Polygons formed by the hypothesized nodes in each corner  (c) New set of nodes optimized with information from the segmentation network (d) Compare to Human annotation: \colorbox{Green}{Green} is human annotation graph \colorbox{Red}{Red} is Pedestrianfer Hypothesized graph. \colorbox{Blue}{Blue} is the optimized graph.}
    \label{fig:sample_opt}
    \vspace{2mm}
\end{wrapfigure}

At each intersection, there are nodes representing sidewalks endpoints and curbs in the hypothesized graph generated by \textit{Pedestrianfer}. Ideally, in a correct pedestrian path network, these nodes would all appear in a connected region segmented as the \textit{corner bulb} class (an example is given in Figure \ref{fig:sample_opt} (c) and  Figure \ref{fig:sample_opt} (d)). To infer the correct geolocation of these nodes, we aim to find a parameterized affine transformation to warp each set of hypothesized nodes in a corner, represented by their pixel coordinates, to a new set of coordinates, so they better align with the \textit{corner bulb} class in the predicted segmentation mask.  We start by connecting the nodes at each corner (shown in Figure \ref{fig:node_hypo}) to form a closed polygon (shown in Figure \ref{fig:node_before}), then we find a parameterized affine transformation so that the sum of the probability of each pixel under each closed polygon being labeled as the \textit{corner bulb} class is the greatest. Mathematically, this optimization process can be defined as follows. For a total of $n$ points in a given image $I$ and each street corner area represented as a polygon $[(x_1, y_1), (x_2, y_2), ... (x_n, y_n)]$, the affine transformation that warps them to a set of new points $[(x'_1, y'_1), (x'_2, y'_2), ... (x'_n, y'_n)]$ can be described as: 

\begin{equation}
\begin{bmatrix}
x'_1 & ... & x'_n\\
y'_1 & ... & y'_n
\end{bmatrix}
= 
\begin{bmatrix}
a & b\\
c & d
\end{bmatrix}
*
\begin{bmatrix}
x_1 & ... & x_n\\
y_1 & ... & y_n
\end{bmatrix}
+
\begin{bmatrix}
t_1\\
t_2
\end{bmatrix}
= AX + t
\end{equation}

That is, $X$ is a set of points representing the coordinates of the nodes in a hypothesized graph at a given street corner. The new set of coordinates 
$[(x'_1, y'_1), (x'_2, y'_2), ... (x'_n, y'_n)]$, and the transformation parameters $A$ and $t$ are to be found using the information obtained from the segmentation network.

Let $m$ be the number of pixels that fall into the polygon enclosed by the $n$ corner points $X$, and let $p_i$ be the probability (as predicted by the segmentation network) that the $i$th pixel is predicted as the \textit{corner bulb} class.  Then we define the function $f$ that finds these pixels and their probability as:

\begin{equation}
f: X  \longmapsto [p_1, p_2, ... p_m]
\end{equation}

,and define a function $g$ that sums $[p_1, p_2, ... p_m]$ as:
\begin{equation}
g(f(AX + t)) = g([p_1, p_2, ... p_m]) = \sum ^m_{i=1} p_i 
\end{equation}

The goal is to maximize $g$ so that the pixels under the new polygons have the greatest sum of probability (shown in Figure \ref{fig:node_after}). Thus, the optimization problem can be expressed as: 

\begin{equation} \label{eq:opt}
\begin{split}
& \underset{A,t}{\text{minimize}}  \quad -g\left(f\left( AX + t \right)\right) \\
& \text{subject to} \quad \forall i \in [1,n] \quad 0 < x'_i < I_{width}, 0< y'_i < I_{height}
\end{split}
\end{equation}

There is no closed-form solution to Equation \ref{eq:opt}, and the gradient of $f$ cannot be explicitly found. Hence we use the simultaneous perturbation stochastic approximation (SPSA) \cite{spall1998implementation} method to find the optimal parameters $A$ and $t$ of the objective function $g$.

After optimizing sidewalk and curb node geolocation, we connect them with the information from the hypothesized graph to generate new sidewalks and crossing edges. The optimization process is shown in Figure \ref{fig:sample_opt}. Figures \ref{fig:node_hypo} and \ref{fig:node_before} show the original node locations in the \textit{Pedestrianfer} hypothesized graph. In this example, \textit{Pedestrianfer} approximates the nodes erroneously in the middle of the road. Figure \ref{fig:node_after} shows the locations of the nodes after optimization, where the nodes and the enclosed polygon formed by the nodes are moved to the middle of a region predicted as the \textit{corner bulb} class. The improved graph (post-optimization) in Figure \ref{fig:node_graph}, shows graph elements considerably closer to the human-generated (ground truth) graph. An evaluation of the overall Prophet system with respect to ground truth appears in Section \nameref{sec:exp}.

\hfill\break%
\hfill\break%
\hfill\break%
\newpage

The optimized graph is used with probability masks of each class from the segmentation network to further improve the predicted path network map accuracy.  For a set of hypothesized nodes at a given corner, if the closed polygon they form does not overlap with enough high-probability pixels representing the \textit{corner bulb class}, we consider these nodes to be falsely hypothesized. Mathematically, we define $\mu_{p}$ as:
\begin{equation}
\mu_{p}= g(p)/m
\label{eq:avg_prob}
\end{equation}
If $\mu_{p}$ is less than a set threshold, these nodes are considered to be falsely hypothesized and therefore directly removed from the graph without optimizing with equation \ref{eq:opt}. The threshold is chosen to balance the edges' precision and recall in this post-processing step. Higher thresholds lead to higher precision.

For downstream applications (e.g. wayfinding) that use the predicted graph, confidence values are added to edges to inform the optimization for high-confidence routes, i.e., the application may choose to avoid low-confidence edges and instead optimize for higher-confidence paths. For each edge created by connecting two nodes in the graph, we assign a confidence value as an attribute of the edge as follows: each \textit{LineString} representing the \textit{crossing} class is converted into a polygon by adding a buffer to each side of the \textit{LineString}, then we compute the mean probability of the pixels in the polygon in the \textit{crossing} class, similar to Equation \ref{eq:avg_prob}. The mean probability is used as the confidence value and stored as an attribute of the edge in the graph. Similarly, the confidence values of the \textit{sidewalk} edges are computed and stored as an attribute of the \textit{sidewalk} edges. The confidence value improves the graph for various transportation network analyses, sidewalk network scoring, personalized routing \cite{bolten2019accessmap}, and other downstream applications.

\section{Skeptic: Collaborative Edits, Validation, and Verification}
\label{sec:edits}
The Skeptic protocol provides a set of instructions and collaborative tooling to engage stakeholders in expert review and validation of the predicted pedestrian pathway graph. While there is a potential role of capacity-building tools --- both technical and socio-cultural tools --- in helping practitioners operationalize and be responsive to accessibility mandates, there is a cautionary tale learned from other domains such as structural engineering and medicine, where tools created without involving practitioners in their design or implementation have been ignored or misused~\cite{mejia2024s}.  Our protocol is informed by several years of practical experience with crowd mappers, transportation planners, and pathway review teams, reified by an interview study conducted in 2022.

We performed sixteen semi-structured interviews with nineteen professionals working at thirteen municipal-, county-, and state-level governments in Maryland, Oregon, and Washington States. Only jurisdictions with existing ADA transition plans were invited to participate. By interviewing agencies who have already engaged to some degree with the needs for accessibility within their pedestrian environments, we hoped to identify participants who understand the challenges of this space and may have developed strategies to address these challenges \cite{eisenberg2020communities}. Interview subjects were identified by requesting access to expertise in the following areas:

\begin{itemize}
    \item Agency goals for improving accessibility, including parts of the planning process
    \item Technical tools and data sources used to make accessibility decisions
    \item Insight into improving the decision-making process regarding accessibility
    \item Roles of those involved in decisions about the implementation of accessibility improvements
    \item Community’s role in decision making
\end{itemize}

We learned that: (1) Experts contribute map data in their agencies only sporadically, and when they do, they focus on specific projects with narrowly scoped areas. The addition of elements in their institutional geographical information systems (GIS) or asset management systems is time-consuming and isolating work. (2) Experts also disclosed that they have been challenged to keep aligned with other institutions, both inside and outside of government. Many interviewees work with the same asset management system as other people in their agency, but due to organizational friction, they do not always work together on planning, communication strategies, and development of resources and tools. 
(3) While many agencies have open, shared public GIS data, the underlying source data is rarely available. At times, even data within the same agency across different departments were not shared. Consequently, professionals doing this work were typically on their own to obtain current, relevant accessible pedestrian data to satisfy the need to measure and account for barriers and assess accessible reach in their jurisdictions. Data maintenance has many stakeholders but few resources devoted to it, a collaborative approach is needed, one which is guided by simple, understandable processes.  As an overarching concern, the interviewees shared the importance of engaging the community.

The Skeptic protocol was designed in response to these findings.  The task-partitioning interface (Section \nameref{sec:provisioning}) rewards rather than penalizes sporadic engagement (Finding (1)) while facilitating mutual validation across agency mappers (Finding (2)).  The collaborative vetting interface (Section \nameref{sec:vetting}) facilitates inter-agency communication and sharing (Finding (3)).  The community validation with those affected by the data (Section \nameref{sec:community}) meets engagement mandates (Finding (4).  



\subsection{Provisioning Editing Tasks}
\label{sec:provisioning}
We identify priority areas for manual validation according to two criteria: diversity (across socioeconomic conditions, zoning, land use, and building standards) and the availability of interested local community participants who volunteer as project stewards.  In our initial pilot, we partnered with the King County Health Through Housing (HTH) initiative, which converts hotels and apartments into low-income housing to combat chronic housing insecurity. HTH was interested in an analysis of access to health services, grocery stores, and schools from HTH sites. In ongoing work, we identify 14 HTH sites and define project regions as a 0.75-mile radius around each site.  In our evaluation, we select 3 of these sites for socioeconomic diversity in the region and the availability of previously expert-mapped data with which to compare.

Each project region is then partitioned into intersection-scale task regions.  A single intersection as the quantum of editing effort allows mappers to complete the work in a few seconds to a few minutes which helps maintain motivation.  Moreover, this unit helps us manage concurrency within the editing tool --- each user locks a task region during their work to prevent concurrent updates.  Finally, these regions ensure principled redundancy --- we can assign the same task to multiple editors to validate inter-mapper agreement. Redundancy also contributes to greater team collaboration and reduced isolation (and therefore fatigue) on the part of contributors.
 

\subsection{Interactive, Collaborative Expert Vetting}
\label{sec:vetting}
We then load the partitioned data into a system to manage interactive concurrent updates. The tasking management software is an open source tool for the Humanitarian OpenStreetMap Team (HOT) \cite{hot} for the coordination of volunteers and organizations to map on OSM. We have created a fork of this open-source project and modified it to align with the findings of our interview study. The HOT tool is publicly available and is backed by a private deployment of the OSM database schema to ensure interoperability with OSM editing tools. This platform-oriented approach facilitates the distribution of tasks across stakeholders. It also provides our team visibility over progress and quality of the mappers' work.


In the pilot study, we recruited two teams and trained them. The UW team consisted of five undergraduate interns who completed a two-hour training on the motivation of the project and the use of the editing tools. The ORISSA team, an industry-based team focused on data annotation, was asked to pilot the tool as a comparison with the UW team. As part of the training, we assigned each team member a project area and 150 intersection-scale tasks within the assigned project, then gave them one week to complete the tasks.  We then reviewed their results, provided feedback, and applied corrections. Each participant was then asked to complete the remainder of the assigned project area within a two-week period. 

While editing, mappers were presented with instructions designed to maximize quality informed by our interview study. The project-agnostic instructions were to 1) edit crossings prior to sidewalks to facilitate connectivity, 2) insert a midpoint in each crossing to represent connectivity with the road, 3) attend to transitions between crossings and sidewalks (e.g., curb ramps), 4) determine whether curb nodes are raised or lowered, and 5) report whether the imagery used to validate the networks is of usable quality. The system also supports project-specific instructions to improve quality; we intend this protocol to be adapted broadly.

After reviewing instructions, each mapper locks a particular task.  If another mapper has already locked that task, the system rejects the lock request and provides an appropriate error message.  If a contributor already has another lock on a different task, they must relinquish that lock before acquiring another.  In this way, we prevent the need to resolve edit conflicts.  Since the task regions are small, lock contention is rare, and the process is largely invisible to mappers.

After editing a task, the mapper comments on the task to provide metadata summarizing (1) the completion status of the task, (2) whether the imagery was of sufficient quality for them to assess the map confidently, and (3) whether they wish to move on to the next task. The lock is relinquished and the edited results are saved to the private OSM database.


%


\subsection{Community Validation "Deep Dives"}
\label{sec:community}
As the mappers complete their work, community volunteers are deployed to perform an extra level of validation, and to enrich the data with locally important information that can only be recorded through direct inspection of the built environment.  For example, the presence or absence of accessible pedestrian signaling (e.g., audible walk signals) is especially important for HTH residents with vision impairments. Although coverage of these features is limited by the scalability issues of direct inspection, the locally enriched data affords locally accessible routing, commensurate analysis of neighborhood-level access to resources and services, and, crucially, a source of training data for future models to automate tagging.

In our pilot, we have recruited teams of HTH residents to conduct mapping efforts around 3 of 14 project areas.  Using a custom application called GoInfoGame (GiG), they complete "quests" to answer questions (i.e., provide tags) in specific task regions.  In contrast to the HOT tool and associated editors, GiG is designed to be dramatically simpler to use --- it does not support geometry editing, it centers a single tag at a time, and can be used from a mobile device.  


\section{Experiments}
\label{sec:exp}
In this section, we first present additional implementation details of Prophet. We then present both qualitative and quantitative evaluation results. 
Since Prophet relies on semantic segmentation as an intermediate step (as is common among state of the art tools~\cite{hosseini2023mapping}), we first provide evaluation on the segmentation models, then provide evaluation of the inferred pedestrian pathway graphs.

\subsection{Training the Segmentation Network}
\label{sec:exp_train}
\begin{figure}[ht!]
    \centering
    \resizebox{0.9\columnwidth}{!}{
    \begin{tabular}{c|c|c|c|c|c|c}
     \textbf{Aerial Image} & \textbf{Rasterized street map} & \textbf{Ground Truth}  & \textbf{Satellite only} & \textbf{Street only} & \textbf{Satellite + Street} & \textbf{Prediction Graph}\\
        \includegraphics[width=0.33\columnwidth]{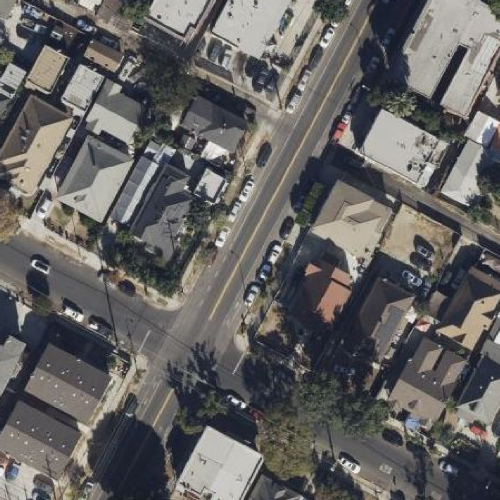}   &
        \includegraphics[width=0.33\columnwidth]{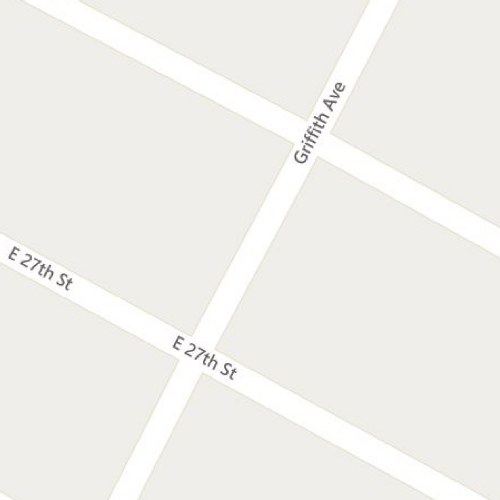} &
         \includegraphics[width=0.33\columnwidth]{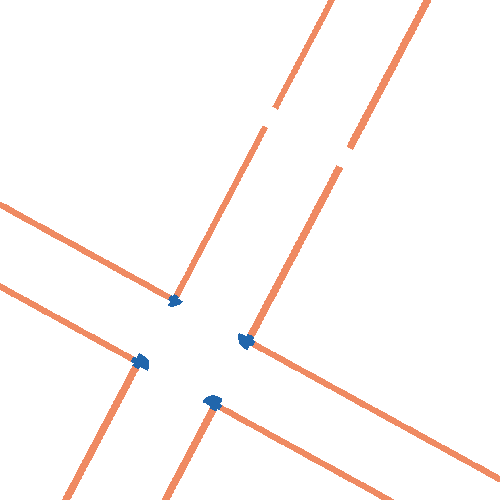} &
        \includegraphics[width=0.33\columnwidth]{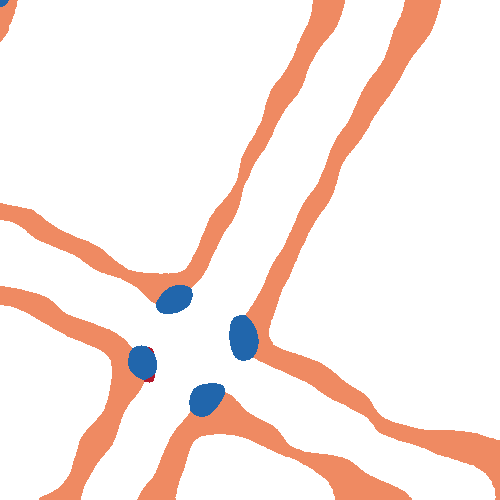} &
        \includegraphics[width=0.33\columnwidth]{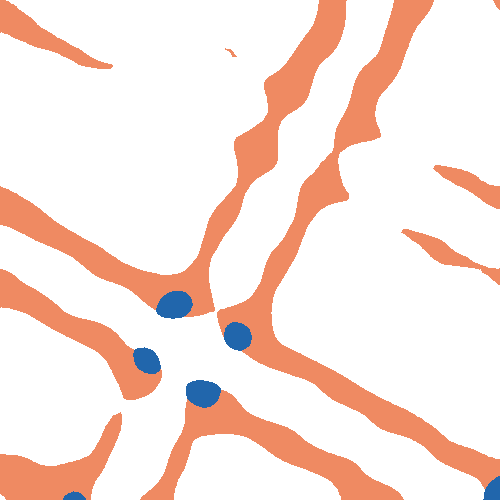} &
        \includegraphics[width=0.33\columnwidth]{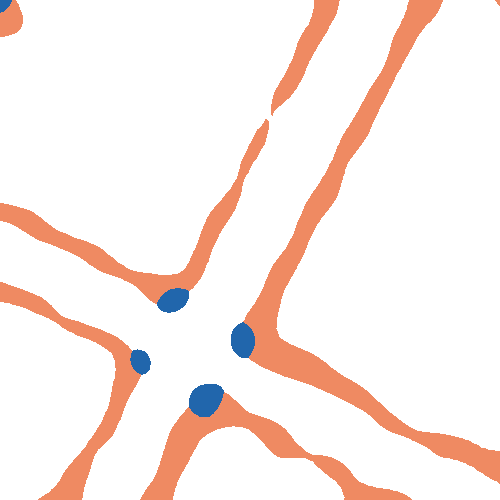} &
        \includegraphics[width=0.33\columnwidth]{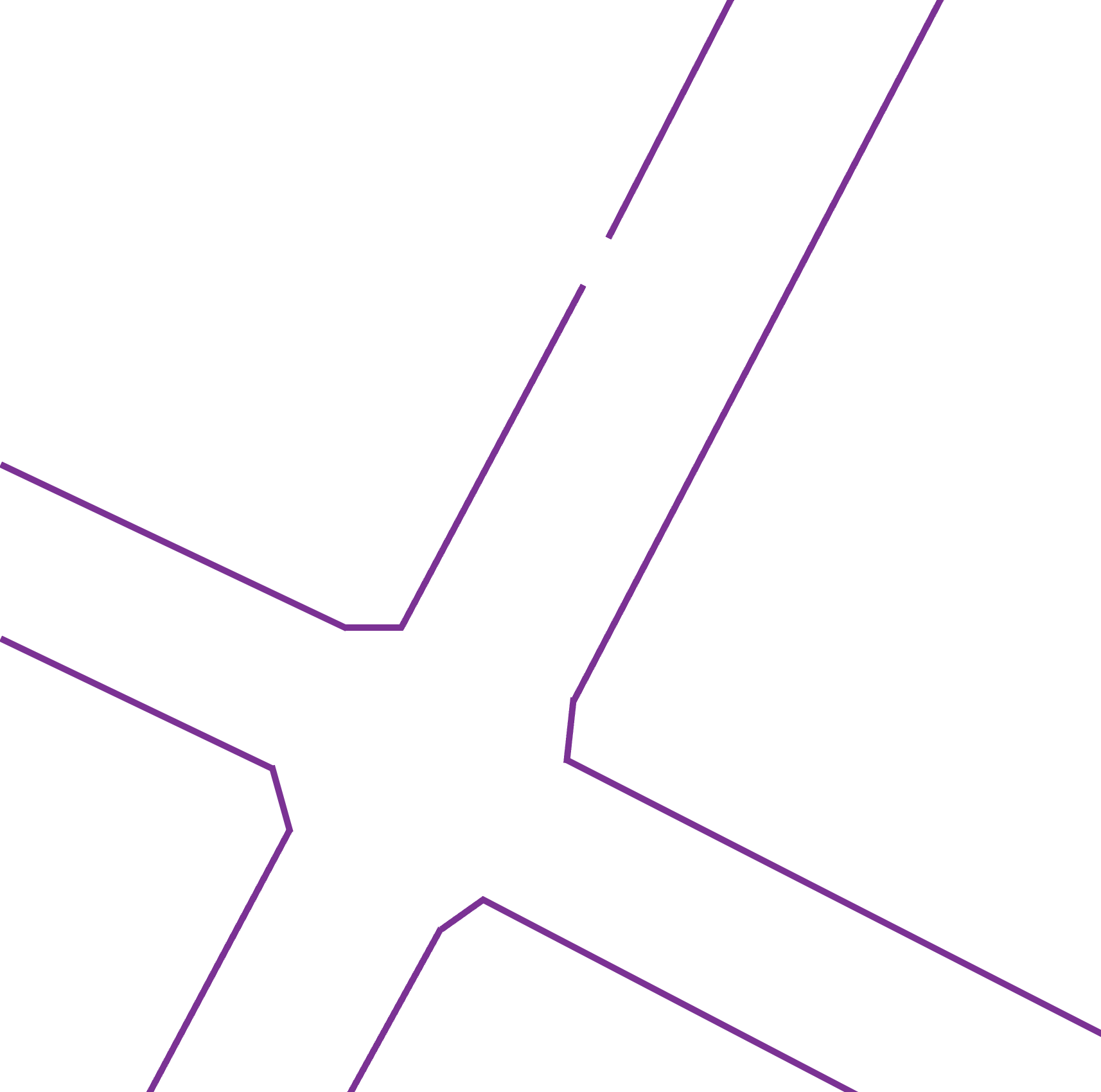} 
        \\
        \includegraphics[width=0.33\columnwidth]{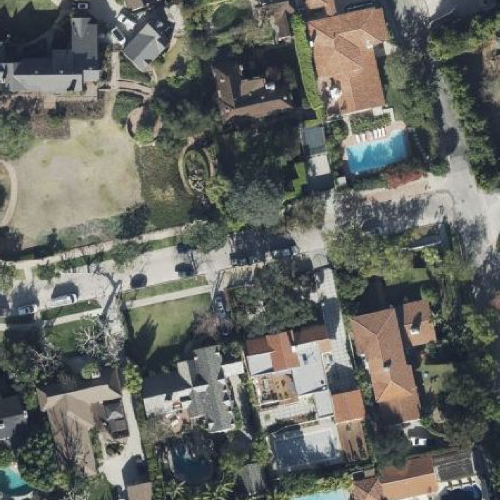}   &
        \includegraphics[width=0.33\columnwidth]{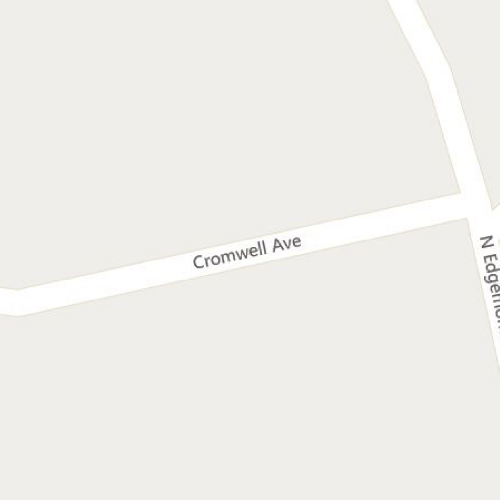} &
         \includegraphics[width=0.33\columnwidth]{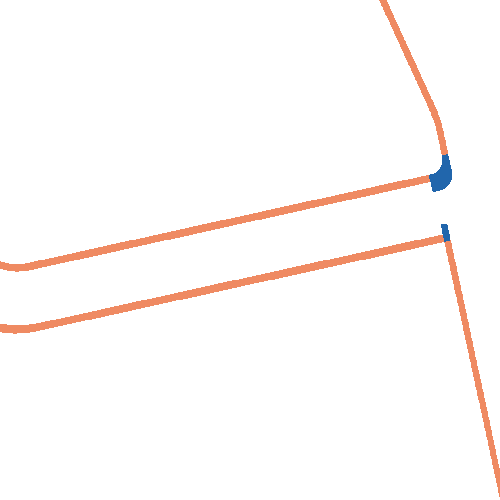} &
        \includegraphics[width=0.33\columnwidth]{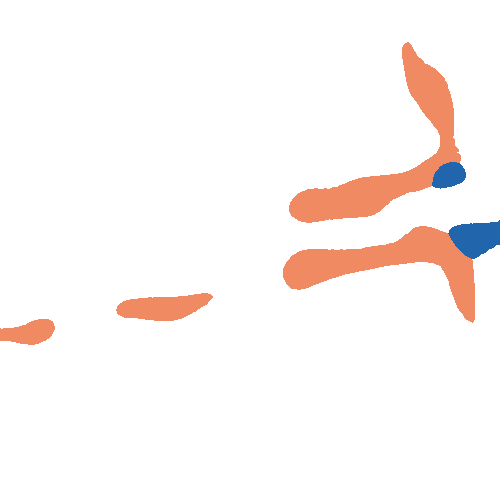} &
        \includegraphics[width=0.33\columnwidth]{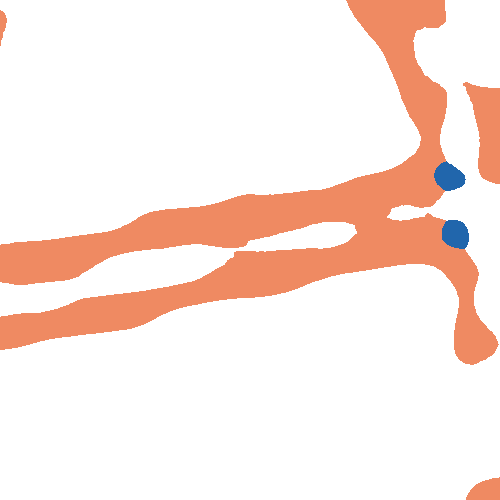} &
        \includegraphics[width=0.33\columnwidth]{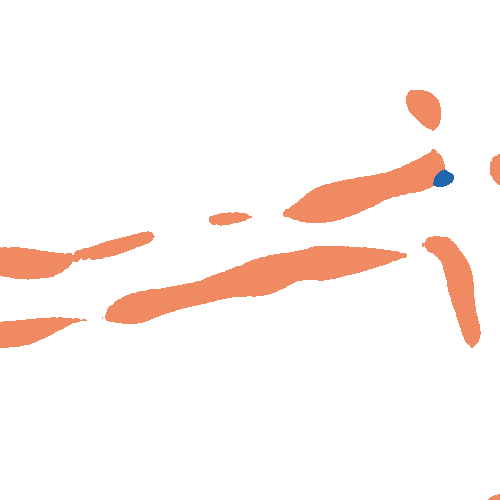} &
        \includegraphics[width=0.33\columnwidth]{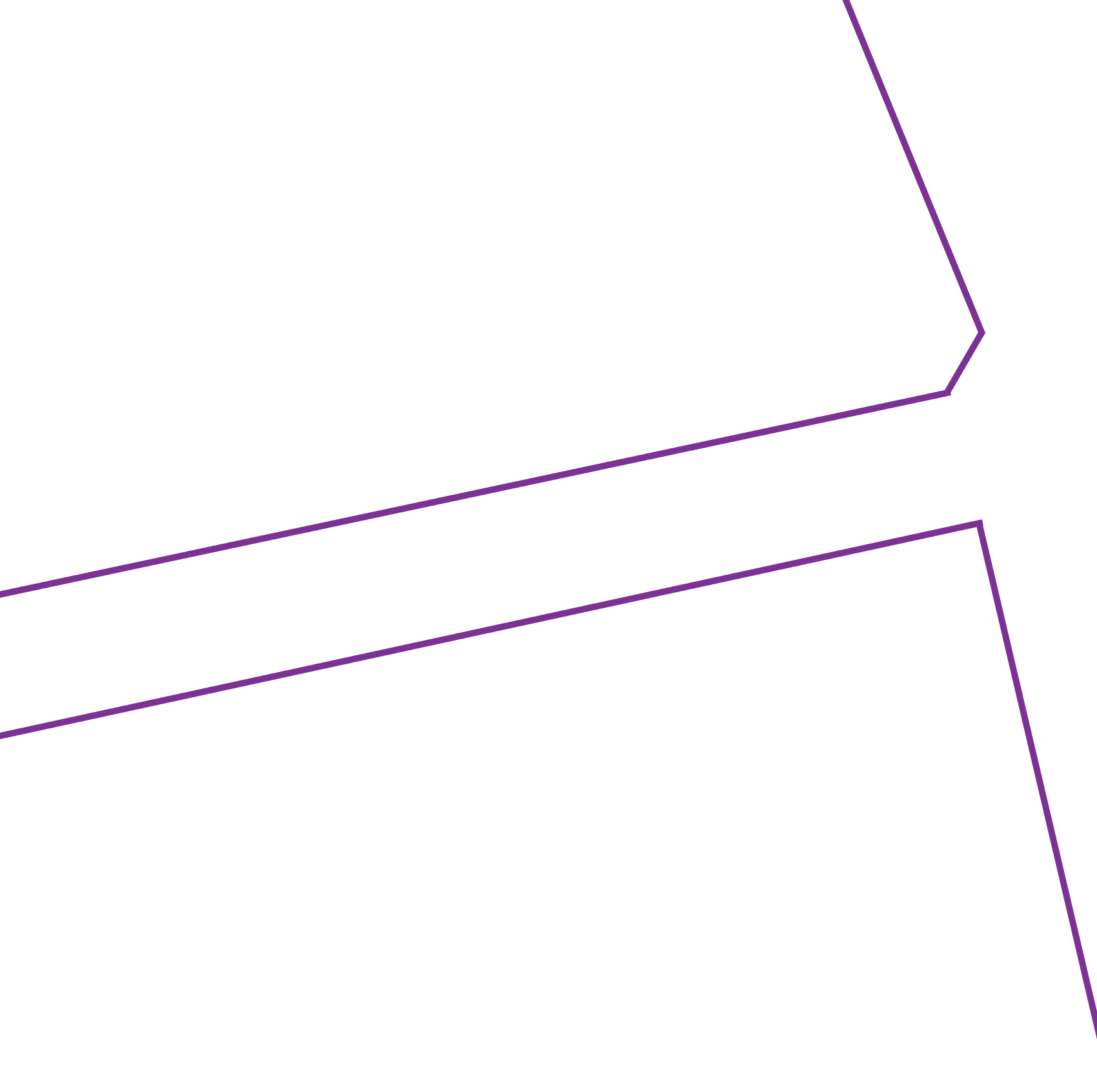} 
        \\
        \includegraphics[width=0.33\columnwidth]{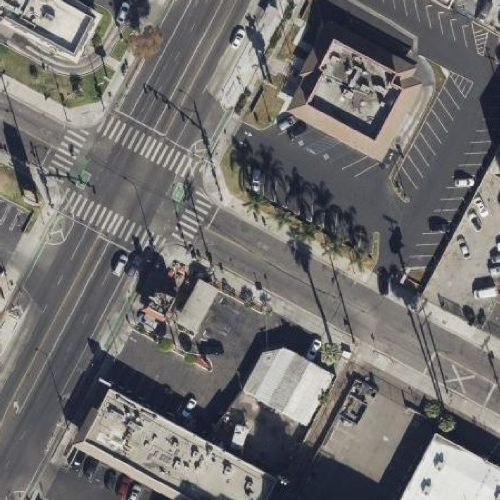}   &
        \includegraphics[width=0.33\columnwidth]{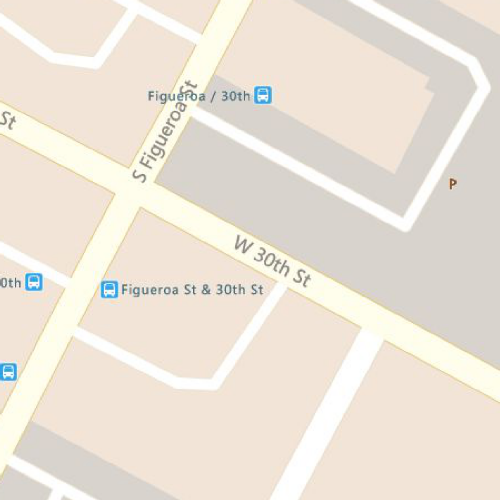} &
         \includegraphics[width=0.33\columnwidth]{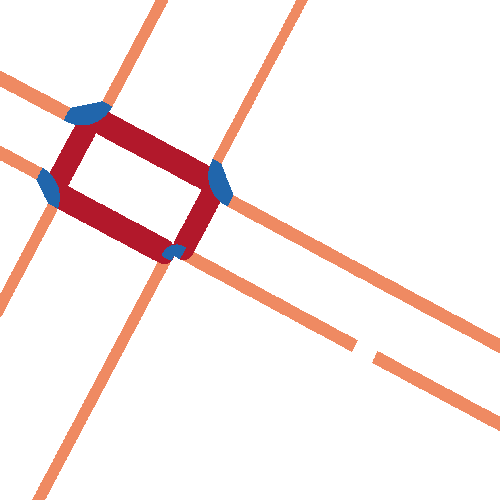} &
        \includegraphics[width=0.33\columnwidth]{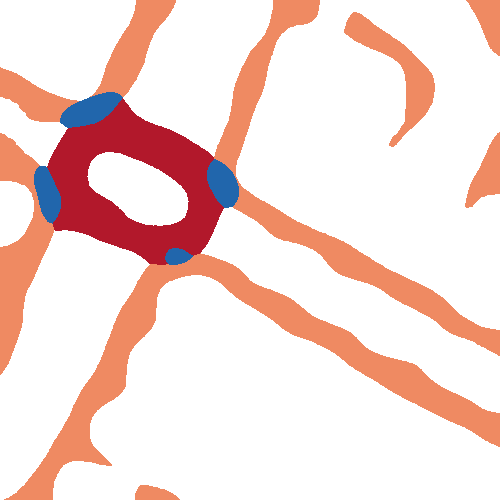} &
        \includegraphics[width=0.33\columnwidth]{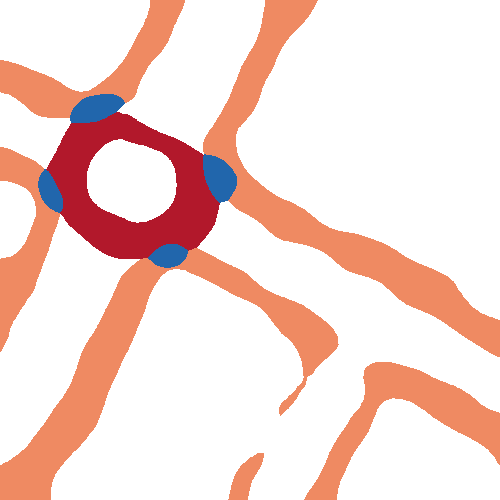} &
        \includegraphics[width=0.33\columnwidth]{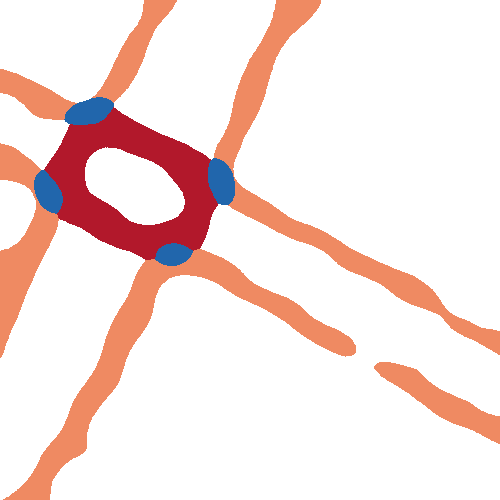} &
        \includegraphics[width=0.33\columnwidth]{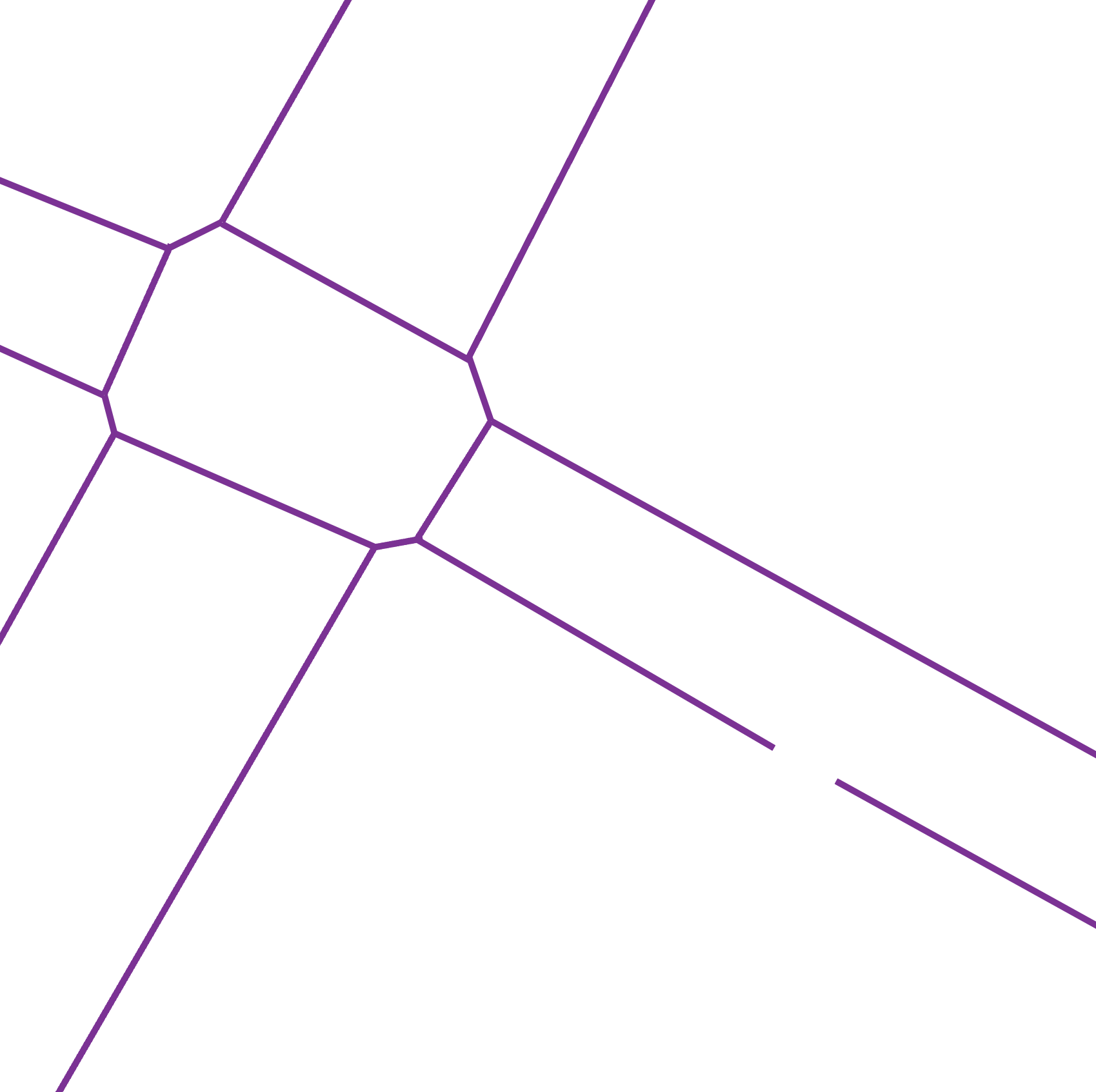} 
        \\
 \end{tabular}
 }
\caption{Qualitative results on the validation set. The segmentation results of 3 different models are shown in columns 4-6. (1) Trained with the aerial satellite image branch only (2) Trained with the street map tile branch only (3) Trained with both the aerial satellite image branch and the street map image tile branch. The segmented classes are \colorbox{corner}{\textcolor{classtext}{\textit{Corner bulb}}}, \colorbox{sidewalk}{\textcolor{classtext}{\textit{Sidewalk}}}, and \colorbox{crossing}{\textcolor{classtext}{\textit{Crossing}}}. The model using both aerial satellite images and street map images outperforms  models that use either input alone. Column 7 illustrates the connected pedestrian pathway graph inferred by Prophet. We discuss these samples and  quantitative results in Section \nameref{sec:exp}.
}
    \label{fig:qual_samples}
\end{figure}

Three segmentation networks were trained with different setups: aerial satellite image branch only, street map image tile branch only, and both aerial and street map image tile branches. All three models use the same dataset split and data augmentation techniques. We train and validate with an $80\%/20\%$ split of the imagery and annotation samples in the PathwayBench dataset \cite{zhang2024pathwaybench}. To improve the robustness of the segmentation network, data augmentations were applied to the training data during the training stage. First, random rotating, cropping, and resizing were applied. In addition, we applied Gaussian kernels with random sizes in training to increase the model's robustness when generating inferences with low-resolution images. The model using both inputs significantly outpeform either source alone; all performance results are shown in Figure \ref{fig:qual_samples} and Table \ref{tab:quan}. 

\subsection{Qualitative Analysis}
\label{sec:model_qual}
Figure \ref{fig:qual_samples} visualizes the segmentation results on the validation set. The segmentation for \textit{sidewalk}, \textit{crossing}, and \textit{corner bulb} align well with the ground-truth segmentation. These qualitative examples also show the difficulty in predicting pedestrian path network classes with single-source input and the improvement gained from adding other input sources.
The prediction made with the street-map-only model 
produced many false-positive \textit{sidewalk} predictions (first row). Including aerial images during model training helped remove these spurious predictions. Predictions made with only aerial images suffered from occlusion due to vegetation, generating sparse, disconnected graphs (second row). Including street image tiles during the training and inference helped recover many occluded sidewalks. We describe the quantitative evaluation of segmentation performance and graph connectivity
in the following sections.

\subsection{Quantitative Segmentation Analysis}
\label{sec:model_quan}
Table \ref{tab:quan} shows the quantitative experiment results for the models trained with three different neural architectures corresponding to the use of different sources: (1)  the aerial satellite image branch only, ignoring the street map imagery, (2) the street map image tile branch only, ignoring the aerial imagery, and (3) both the aerial satellite images and street map images with a siamese-like network structure. The metrics we included are (1) overall mIoU (mean intersection over union of the predicted and ground truth regions), (2) the mIoU for each of the four classes (\textit{background}, \textit{sidewalk}, \textit{corner bulb}, and, \textit{crossing}) in the dataset, and (3) pixel accuracy.

\begin{minipage}[t!]{0.95\textwidth}
    \vspace{2mm}
    \centering
    \captionsetup{width=0.99\textwidth}
    \captionof{table}{Quantitative segmentation results: the model that uses both aerial satellite images and street map tiles outperforms models that use only one branch of data}
    \label{tab:quan}
    \resizebox{\textwidth}{!}{%
        \begin{tabular}{l|cccccc}
        \toprule
        {\textbf{Method}}
        &  \textbf{Background IoU} & \textbf{Sidewalk IoU} & \textbf{Corner IoU} & \textbf{Crossing IoU} & \textbf{mIoU} & \textbf{Pixel Accuracy} \\
        \midrule
        \textbf{Satellite Only (VGG-16)} & 0.79 & 0.42 & 0.45 & 0.42 & 0.55 & 0.67\\
        \textbf{Street Only (VGG-16)} & 0.75 & 0.33 & 0.41 & 0.37 & 0.51 & 0.66\\
        \textbf{Satellite + Street (VGG-16)} & 0.78 & 0.46 & 0.45 & 0.47 & \textbf{0.57} & 0.69\\
        \textbf{Satellite Only (DeepLabv3)} & 0.77 & 0.57 & 0.47 & 0.51 & 0.63 & 0.74\\
        \textbf{Street Only (DeepLabv3)} & 0.75 & 0.56 & 0.43 & 0.51 & 0.61 & 0.72\\
        \textbf{Satellite + Street (DeepLabv3)} & 0.85 & 0.62 & 0.47 & 0.57 & \textbf{0.67} & 0.76\\
        \textbf{Satellite Only (ViT)} & 0.78 & 0.59 & 0.48 & 0.53 & 0.64 & 0.75\\
        \textbf{Street Only (ViT)} & 0.75 & 0.56 & 0.43 & 0.51 & 0.62 & 0.73\\
        \textbf{Satellite + Street (ViT)} & 0.84 & 0.63 & 0.48 & 0.59 & \textbf{0.69} & 0.78\\
        \bottomrule
        \end{tabular}
    }
    \vspace{2mm}
\end{minipage}\hfill

The segmentation backbones we tested include (1) VGG-16~\cite{simonyan2014very} (2) DeepLabv3 (ResNet-50)~\cite{chen2017rethinking}, and (3) ViT-Base~\cite{dosoViTskiy2020}; our contribution is compatible with any neural vision model.  We observe that the model trained with both aerial and street map imagery outperforms single-source models on every class.

\subsection{Quantitative Evaluation of the Inferred Pathway Graphs}
\label{sec:eval_graph}
Pixel-wise measures do not reflect the routability of the predicted graph. mIoU (or other pixel-wise measures) cannot measure how close a predicted graph is to the ground truth graph. To measure routability, we use graph-level metrics  proposed in previous work \cite{zhang2024pathwaybench}. Three categories of metrics are considered (1) local, polygon-level routability metrics including average number of connected components (avg CC), the average betweenness centrality (avg BC), and \textit{TraversabilitySimilarity}, (2) count-based metrics (node count, edge count, and avg degree for the entire test area), and (3) the F1 scored based on the network segment edge-retrieval method described in previous work \cite{hosseini2023mapping}. 

TraversabilitySimilarity is specialize metric designed to provide a \textit{local} estimate of \textit{global} routability and enable comparisons across predicted networks from different sources~\cite{zhang2024pathwaybench}. We first identify center points of intersections using OSM data. 
We then construct a Voronoi diagram to partition the space into polygons. For each polygon, we record the pairs of polygon boundaries for which a given path network affords \textit{traversal}: a pair of boundaries is traversable if the underlying graph provides a connected path between them.
If the ground truth network (in our case, the graph provided by independent expert mappers) affords traversal between a pair of boundaries, then a predicted graph should as well.  For each polygon, the Jaccard similarity (size of the intersection divided by the size of the union) is computed between the predicted set of traversable pairs and the ground truth set of traversable pairs; the mean of these score is the overall TraversabilitySimilarity. This metric rewards predictions that reproduce connected paths for traversing an intersection, while tolerating local variations in geometry and graph structure (e.g., modeling a curved path with 8 segments instead of 5 has little effect on downstream routing but can severely confound edge-counting metrics.)

\begin{minipage}[htp!]{0.92\textwidth}
    \centering
    \captionsetup{width=\textwidth}
    \captionof{table}{Pathway Network Graph Routability Evaluation}
    \label{tab:graph_eval}
    \resizebox{0.99\textwidth}{!}{%
    \begin{tabular}{l|l|ccc|cc|cc}
    \toprule
    \multirow{2}{*}{\textbf{Method}} & \multirow{2}{*}{\textbf{Area}} & \multicolumn{3}{c|}{\textbf{Global}} & \multicolumn{2}{c|}{\textbf{Local}} & \multicolumn{2}{c}{\textbf{Local (relative to Ground Truth)}}\\
    & & $\textbf{\# nodes}$ & $\textbf{\# edges}$ & $\textbf{avg degree}$ & $\textbf{avg CC}$ & $\textbf{avg BC}$ & $\textbf{edge-retrieval F1}$ & $\textbf{\textit{TraversabilitySimilarity}}$\\ 
    \midrule
    Ground & Argyle & 7576 & 8331 & 0.17 &  1.92 & 0.12 & 1.0 & 1.0\\ 
    Truth    & Bertha Pitts  & 6994 & 7693 & 0.17 & 2.07 & 0.11 & 1.0  & 1.0\\
    & Burbridge  & 3948 & 4408 & 0.20 & 1.82 & 0.12 & 1.0  & 1.0\\
    \midrule
     & Argyle & 8386 & 7872 & 0.18 & 3.44 & 0.05 &  0.98 & 0.32\\ 
     Tile2net & Bertha Pitts & 7744 & 7274 &  0.16 & 3.60 &  0.04 &  0.98 & 0.39\\
     & Burbridge & 4349  & 3954 & 0.24 & 3.13 &  0.04 & 0.98 & 0.31\\ 
    \midrule
     & Argyle & 3623 & 3932 & 0.32 & 1.40 & 0.11 &  0.97 & 0.47\\ 
     Prophet & Bertha Pitts & 2964 & 3169 & 0.34 & 1.25 & 0.10 &  0.97 & 0.44\\
         & Burbridge & 2383 & 2630 & 0.26 & 0.88 & 0.13 & 0.96  &  0.31\\ 
     \midrule
     Prophet & Argyle & 4399 & 5019 & 0.12 & 1.76 & 0.11 & 0.99 & 0.84\\ 
         + Skeptic & Bertha Pitts & 3424 & 3885 & 0.19 & 1.87 & 0.10 & 0.99 & 0.88\\
         & Burbridge & 3137 & 3489 & 0.19 & 1.85 & 0.11 & 0.99 & 0.84\\ 
    \bottomrule
    \end{tabular}
    }
    \vspace{2mm}
\end{minipage}

\begin{figure}[htbp] 
    \centering
    \resizebox{\columnwidth}{!}{
        \begin{tabular}{cccc}
             & \textbf{Tile2net} & \textbf{Prophet} & \textbf{Prophet + SKEPTIC} \\
             & \raisebox{0.9\height}{\rotatebox{90}{\textbf{Argyle}}}
             \includegraphics[width=0.33\columnwidth]{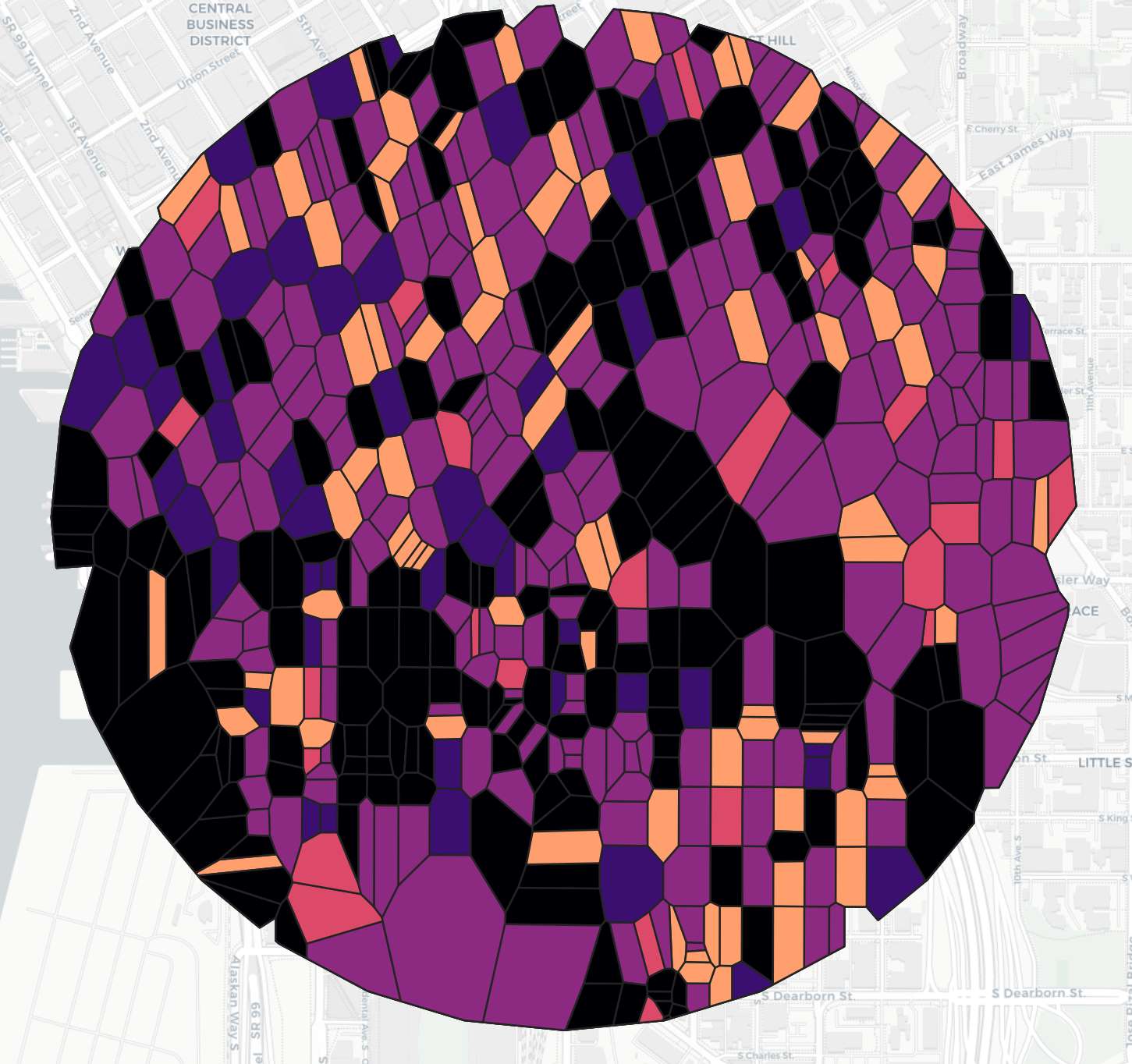} & 
             \includegraphics[width=0.33\columnwidth]{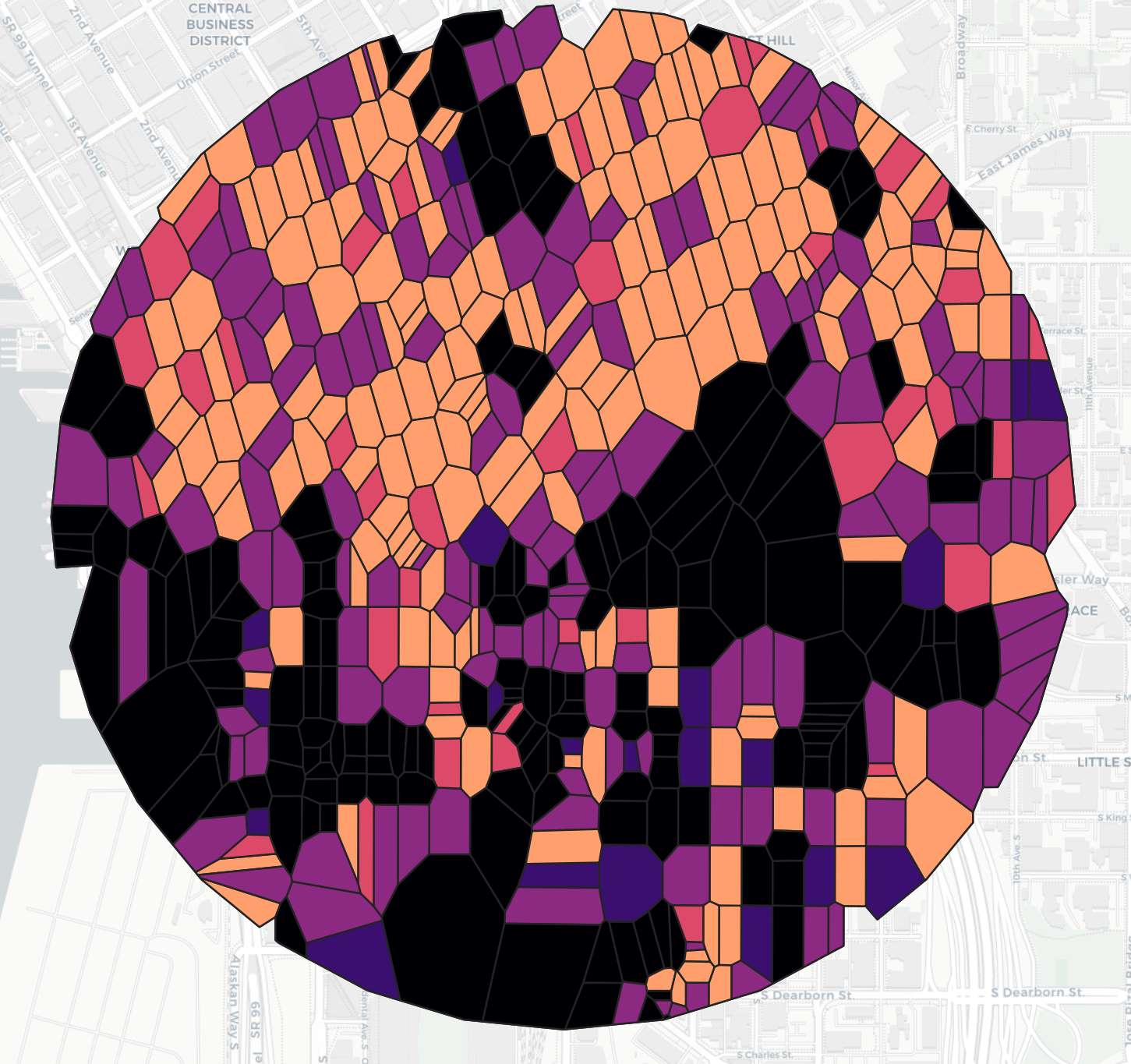} & 
             \includegraphics[width=0.33\columnwidth]{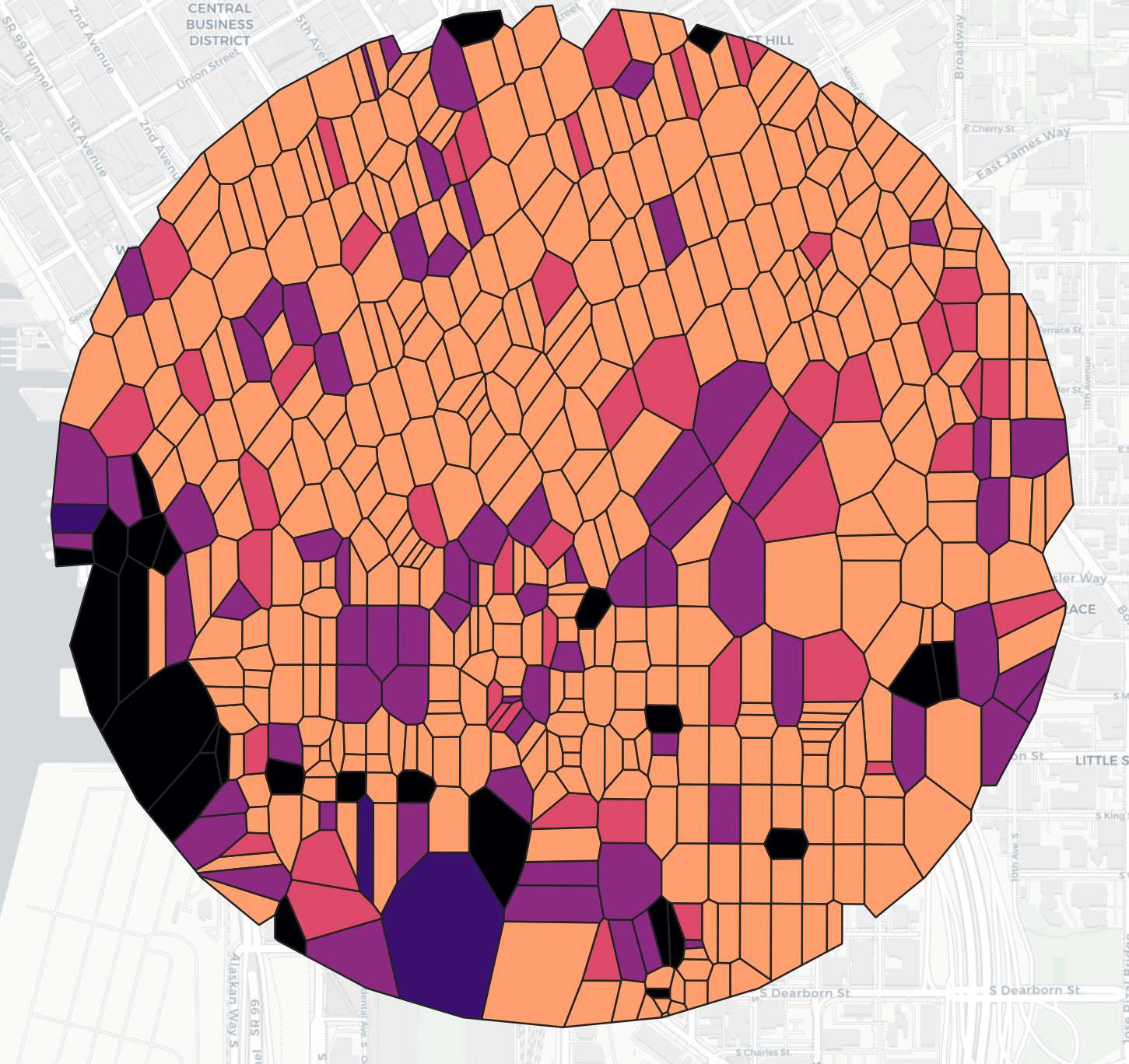} \\

              & \raisebox{0.7\height}{\rotatebox{90}{\textbf{Bertha Pitts}}}
             \includegraphics[width=0.33\columnwidth]{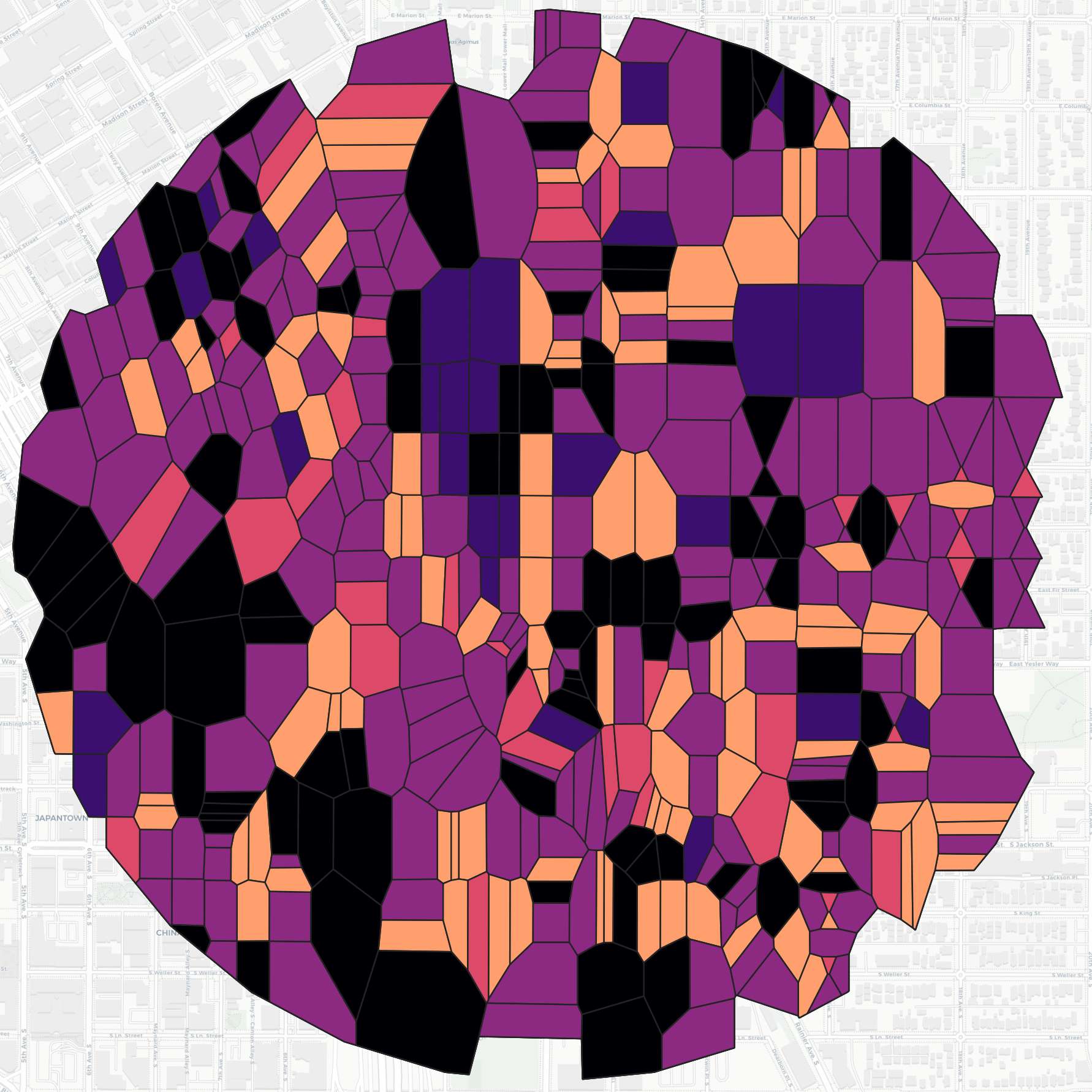} & 
             \includegraphics[width=0.33\columnwidth]{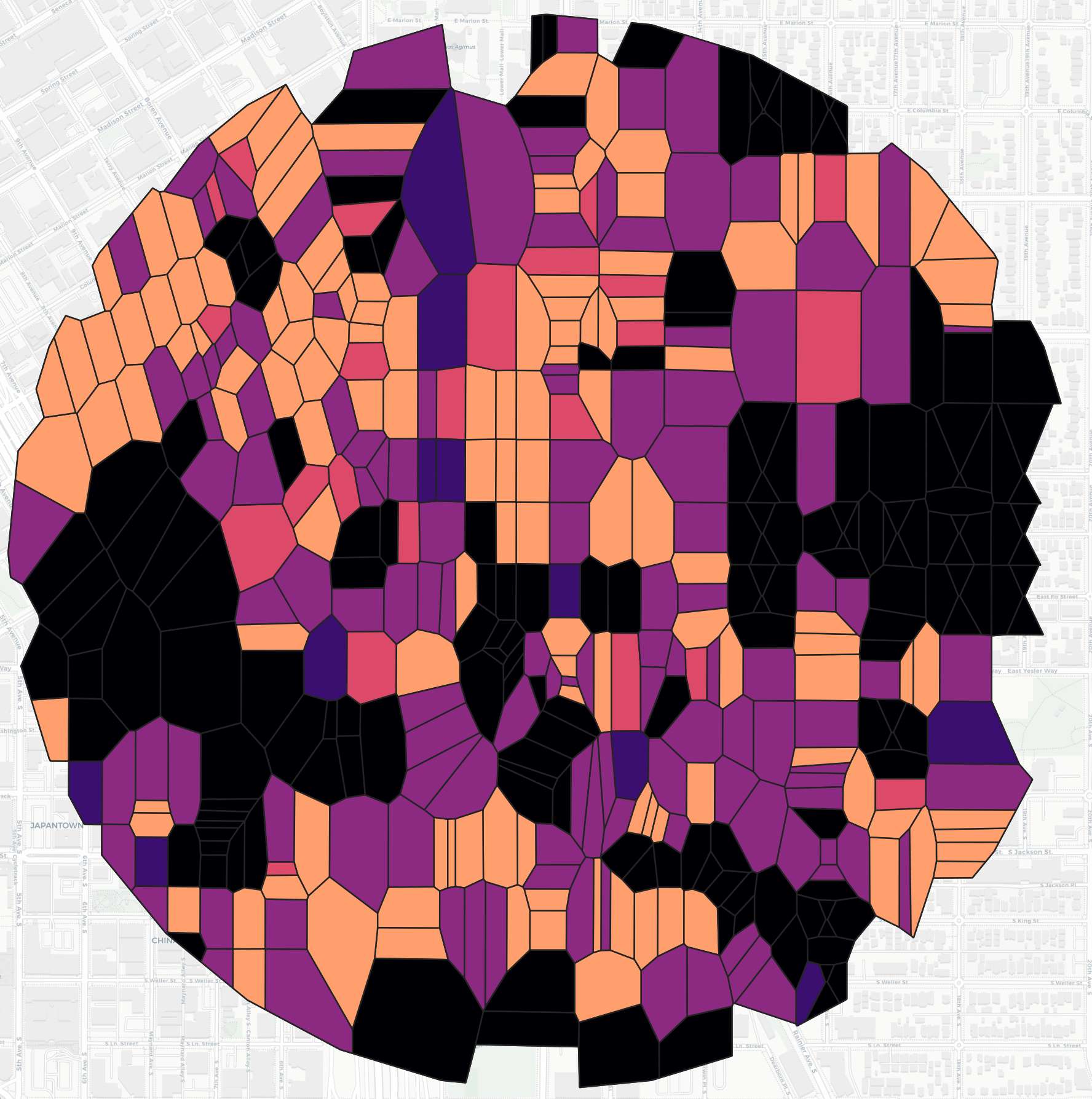} & 
             \includegraphics[width=0.33\columnwidth]{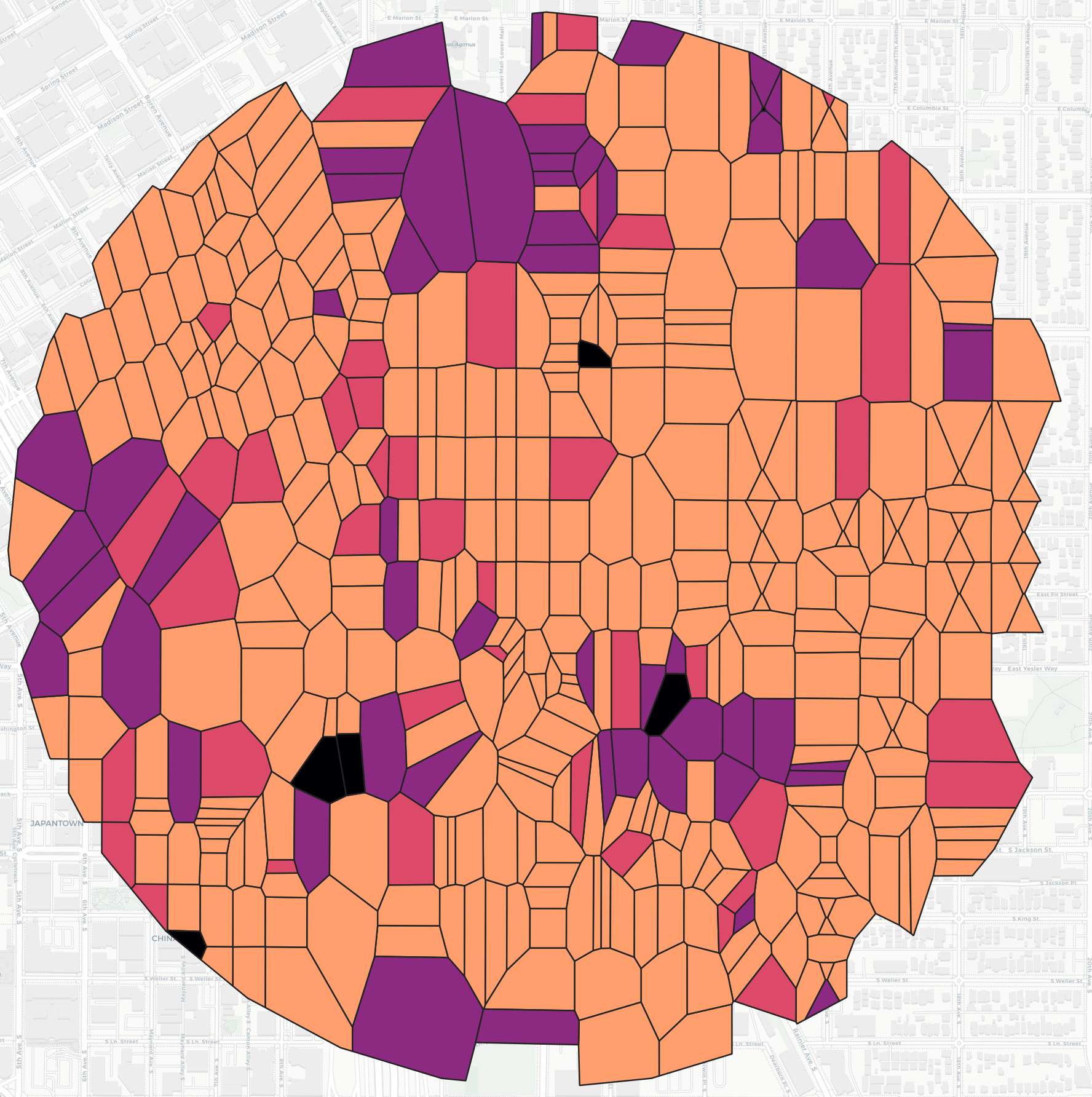} \\

             & \raisebox{0.9\height}{\rotatebox{90}{\textbf{Burbridge}}}
             \includegraphics[width=0.33\columnwidth]{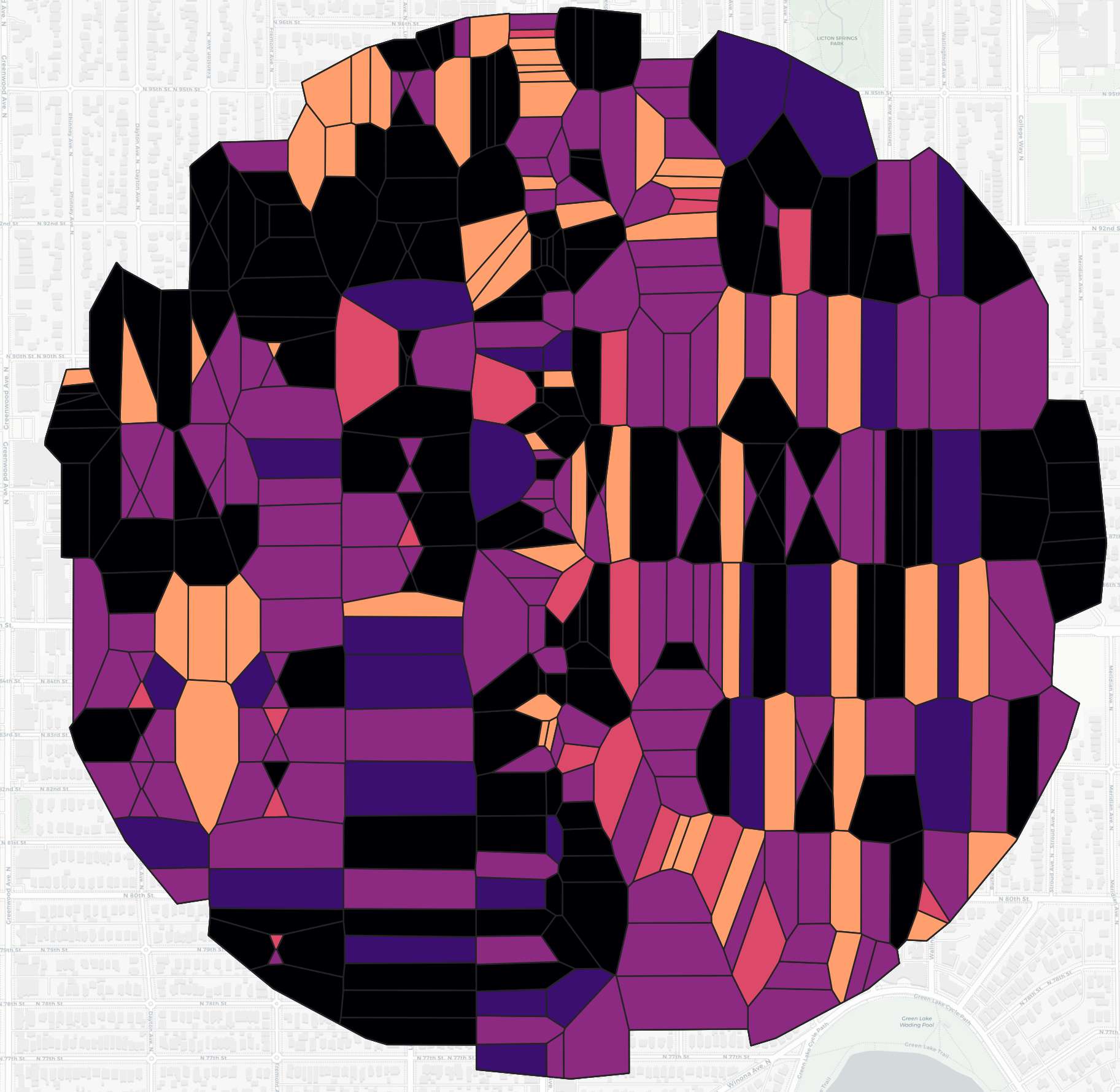} & 
             \includegraphics[width=0.33\columnwidth]{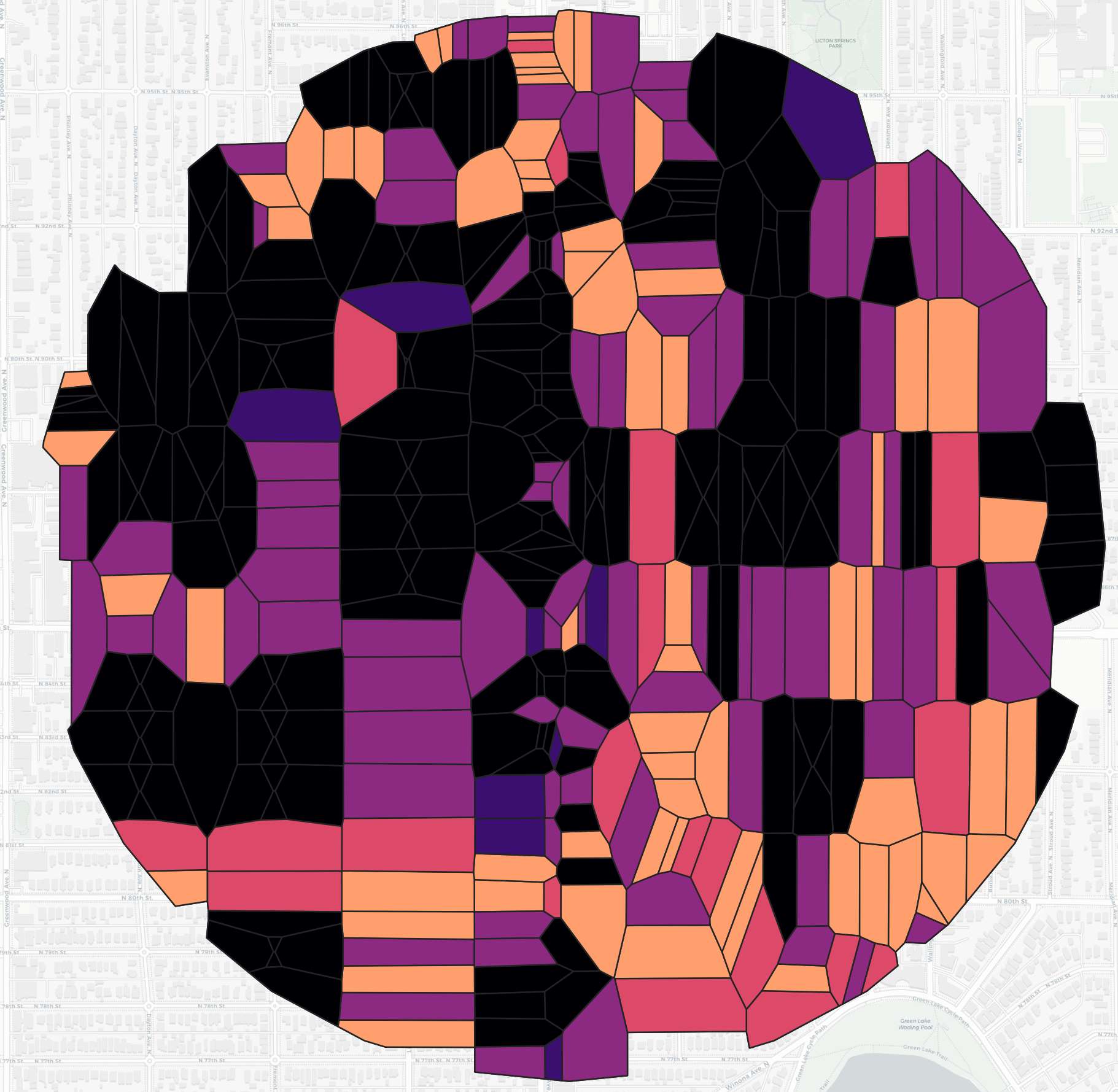} & 
             \includegraphics[width=0.33\columnwidth]{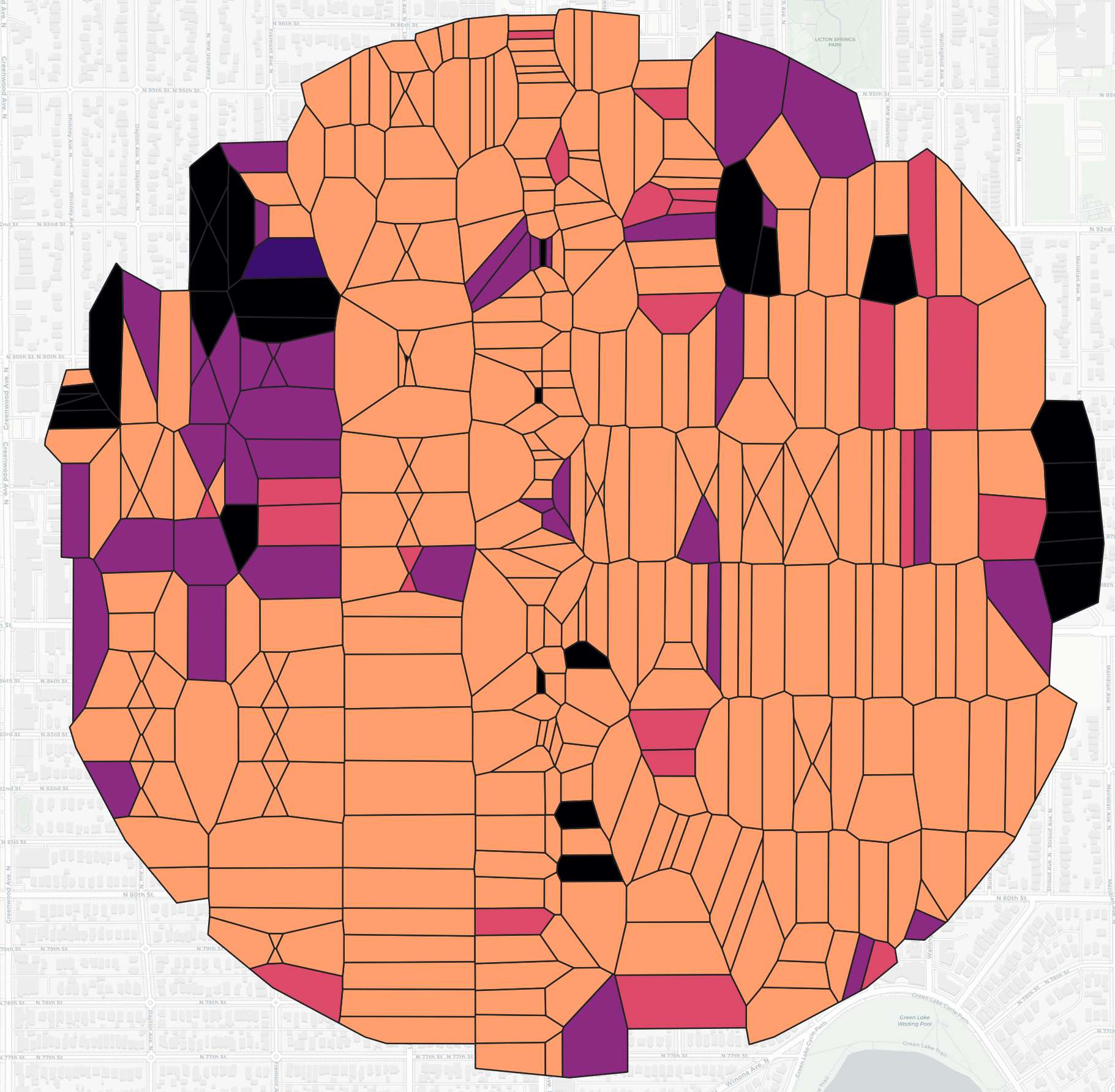} \\
        \end{tabular}
        
    }
    \caption{TraversabilitySimilarity Scores of pathway graphs produced by three methods on three distinct test areas. Brighter colors of each polygon indicate a higher \textit{TraversabilitySimilarity}, while darker colors indicate a lower \textit{TraversabilitySimilarity}. Prophet can generate graphs at scale which better capture the true traversability, than the graphs genearated by the curent state-of-the-art method Tile2Net. Human editing with Prophet + Skeptic produce graphs that best capture the true traversability.}
    \label{fig:ts}
\end{figure}

Evaluation results are summarized in Table \ref{tab:graph_eval}. The metrics in the 'Global' and 'Local' columns indicate whether the property applies to the overall graph (global) or the average of a polygon-local property (local). 
A predicted graph with values closer to the Ground Truth indicates better model performance. 
The 'Edge-retrieval F1' and 'TraversabilitySimilarity' metrics provide direct comparisons to the Ground Truth, where higher scores indicate better model performance. In particular, \textit{TraversabilitySimilarity} compares whether local intersections afford the same ingress and egress points in both predicted and ground truth graphs. Avg CC and avg BC provides additional measures of local connectivity, where the distance from ground truth values exposes deviations from the ground truth graph structure. 

Three methods are included in the evaluation: (1) \textbf{Tile2Net} \cite{hosseini2023mapping}: This previous work uses aerial satellite images to segment sidewalks, crosswalks, and footpaths in cities. It then simplifies the segmented polygons, and extracts the centerlines of the polygons to represent pathway graphs. (2) \textbf{Prophet}: Our method for generating pathway graphs, using both the aerial satellite images and street map image as input and the ViT-Base \cite{dosoViTskiy2020} as backbone in the segmentation model. (3) \textbf{Prophet + Skeptic}: Facilitated human editing applied to Prophet outputs.

As shown in Table \ref{tab:graph_eval}, Tile2Net, despite robust results in terms of the number of edges and nodes, and relatively high edge-retrieval F1 scores across all test areas, shows considerable difference in avg CC and avg BC values across all test areas when compared to the ground truth. Tile2Net tends to overpredict disconnected edges, generating more connected components and lower average betweenness centrality. Moreover, the \textit{TraversabilitySimilarity} scores of Tile2Net are low, suggesting that global paths using the Tile2Net graph would involve significant deviations from ground truth, i.e., they would tend to route pedestrians through completely different intersections. Conversely, Prophet shows better performance in graph routability. They achieve higher \textit{TraversabilitySimilarity} values than Tile2net on the test areas Argyle and Bertha Pitts, despite the F1 scores being sightly lower than those of Tile2Net. Since the F1 score captures similarity between predicted and ground truth graphs in terms of single edge's accuracy, it does not reflect how travelers navigate the environment. The F1 score is therefore less useful for model evaluation and model selection in downstream applications that rely on routing. On the other hand, \textit{TraversabilitySimilarity} better aligns with the practical requirements of real-world routing and navigation tasks. A higher \textit{TraversabilitySimilarity} score of a model indicates a more accurate representation of the true connectivity and routability in its generated graph, as compared to the Ground Truth graph. Prophet outperforms the state-of-the-art method (Tile2Net) in terms of generating pedestrian pathway graphs that capture the true connectivity and traversability of the built environment. 

With the addition of Skeptic, mappers can start from the pathway graphs generated by Prophet, and make edits as needed effificiently (because most of the graph components are alreayd populated by Prophet). Using Prophet and Skeptic together generates graphs that accurately capture the true connectivity and routability in the built environment (above 0.8 TraversabilitySimilarity score in all test areas, per-polygon scores illustrated in Figure \ref{fig:ts}).

\subsection{Generating Pedestrian Pathway Graphs at Scale with Prophet}
\begin{figure}[htbp!]
    \centering
        \begin{subfigure}[b]{0.31\columnwidth}
            \centering
            \includegraphics[width=\columnwidth, height=90px]{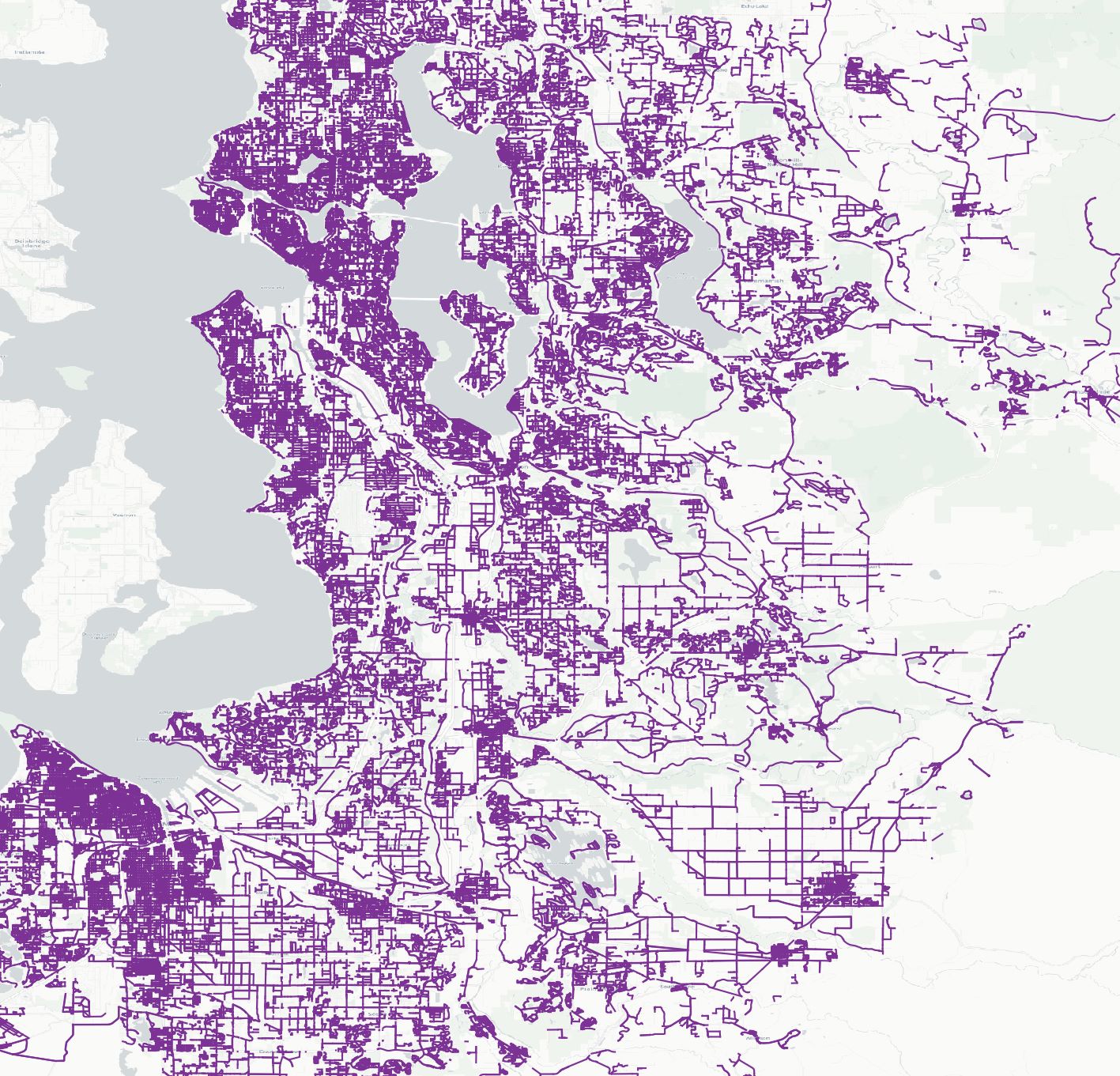}
            \caption{King County, WA}
        \end{subfigure}
        \hfill
        \begin{subfigure}[b]{0.31\columnwidth}
            \centering
            \includegraphics[width=\columnwidth, height=90px]{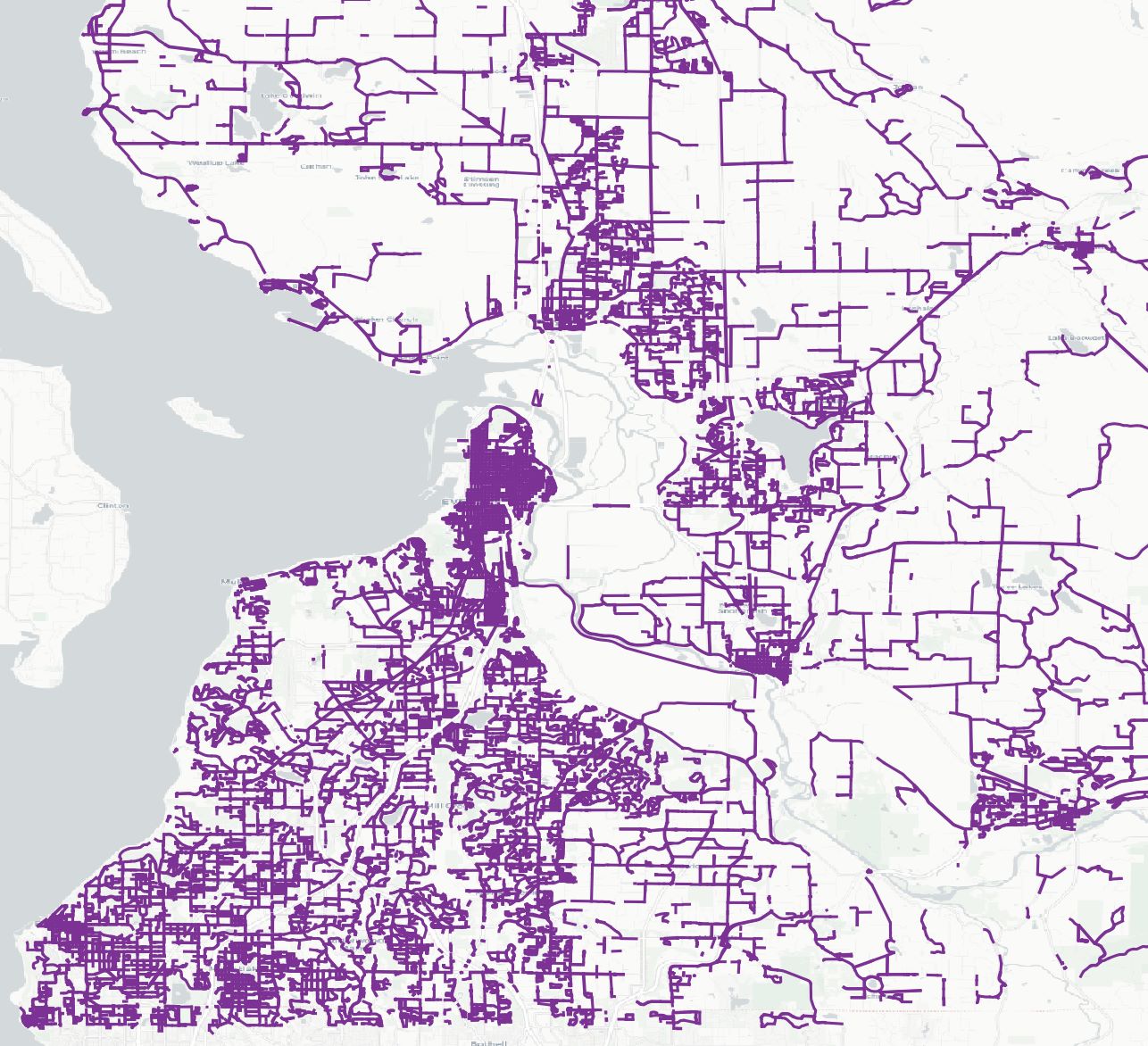}
            \caption{Snohomish County, WA}
        \end{subfigure}
        \hfill
        \begin{subfigure}[b]{0.31\columnwidth}
            \centering
            \includegraphics[width=\columnwidth, height=90px]{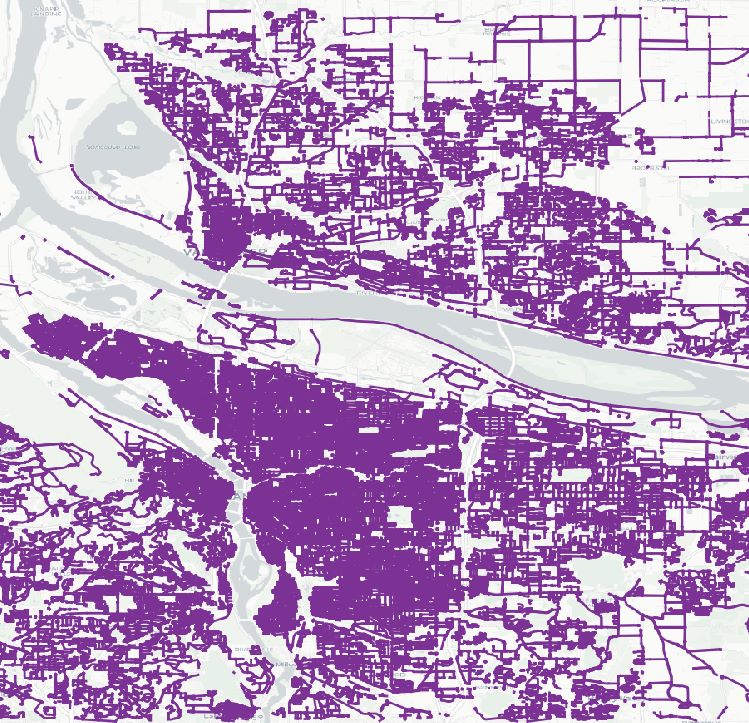}
            \caption{Multnomah County, OR}
        \end{subfigure}
                \begin{subfigure}[b]{0.31\columnwidth}
            \centering
            \includegraphics[width=\columnwidth, height=90px]{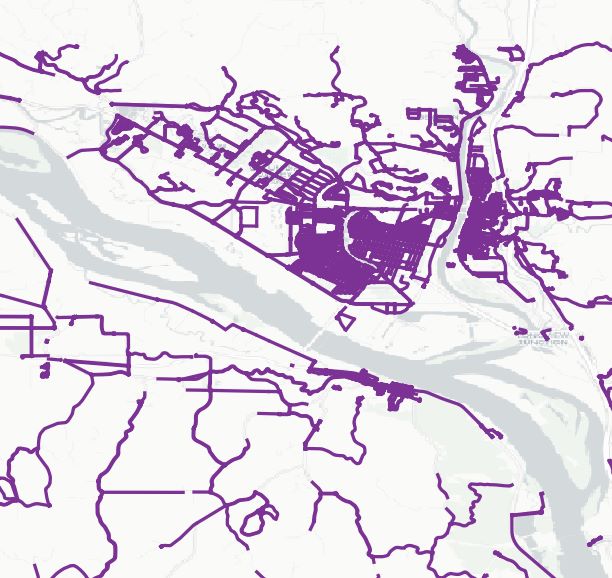}
            \caption{Columbia County, OR}
        \end{subfigure}
        \hfill
        \begin{subfigure}[b]{0.31\columnwidth}
            \centering
            \includegraphics[width=\columnwidth, height=90px]{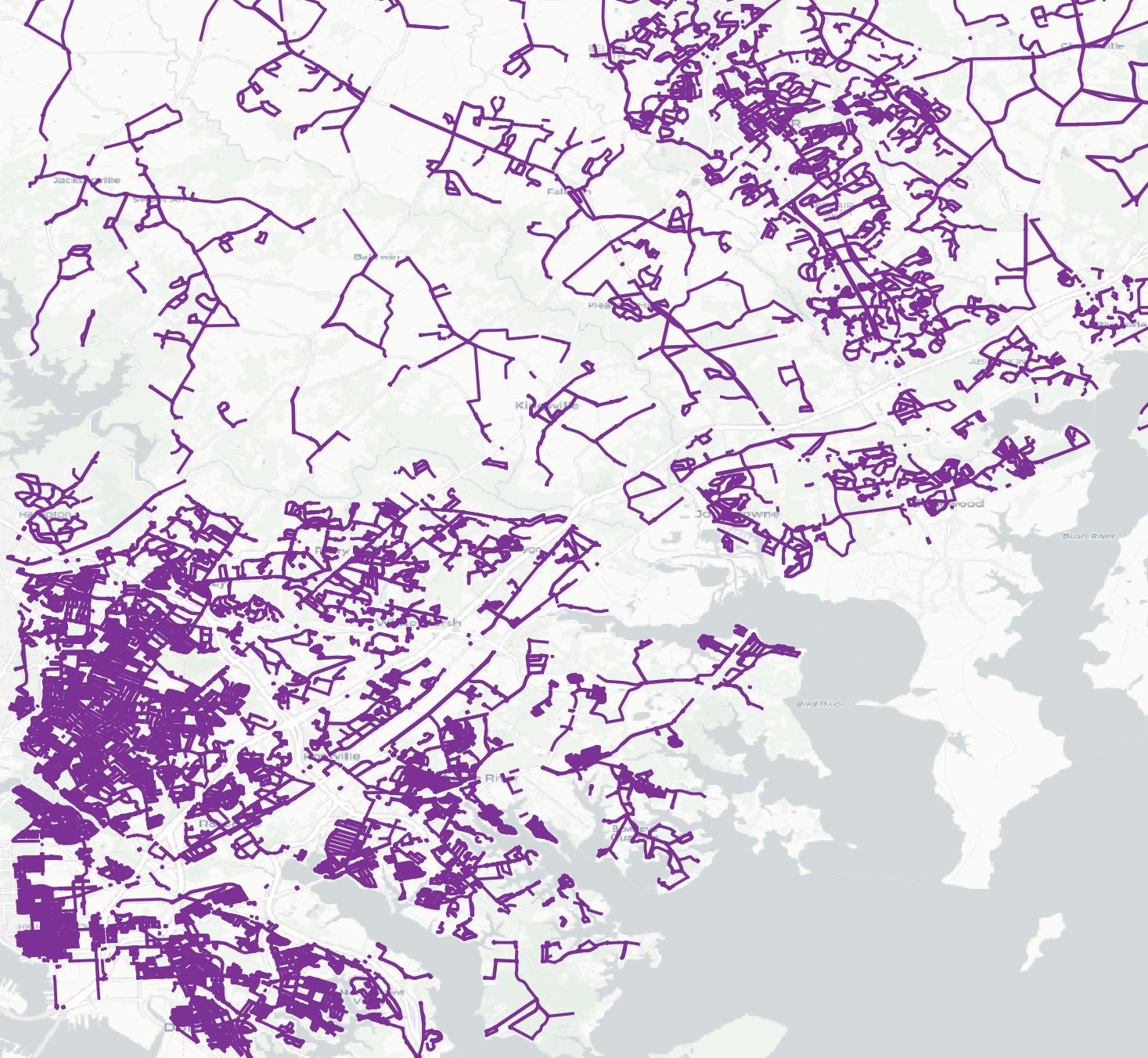}
            \caption{Harford County, MD}
        \end{subfigure}
        \hfill
        \begin{subfigure}[b]{0.31\columnwidth}
            \centering
            \includegraphics[width=\columnwidth, height=90px]{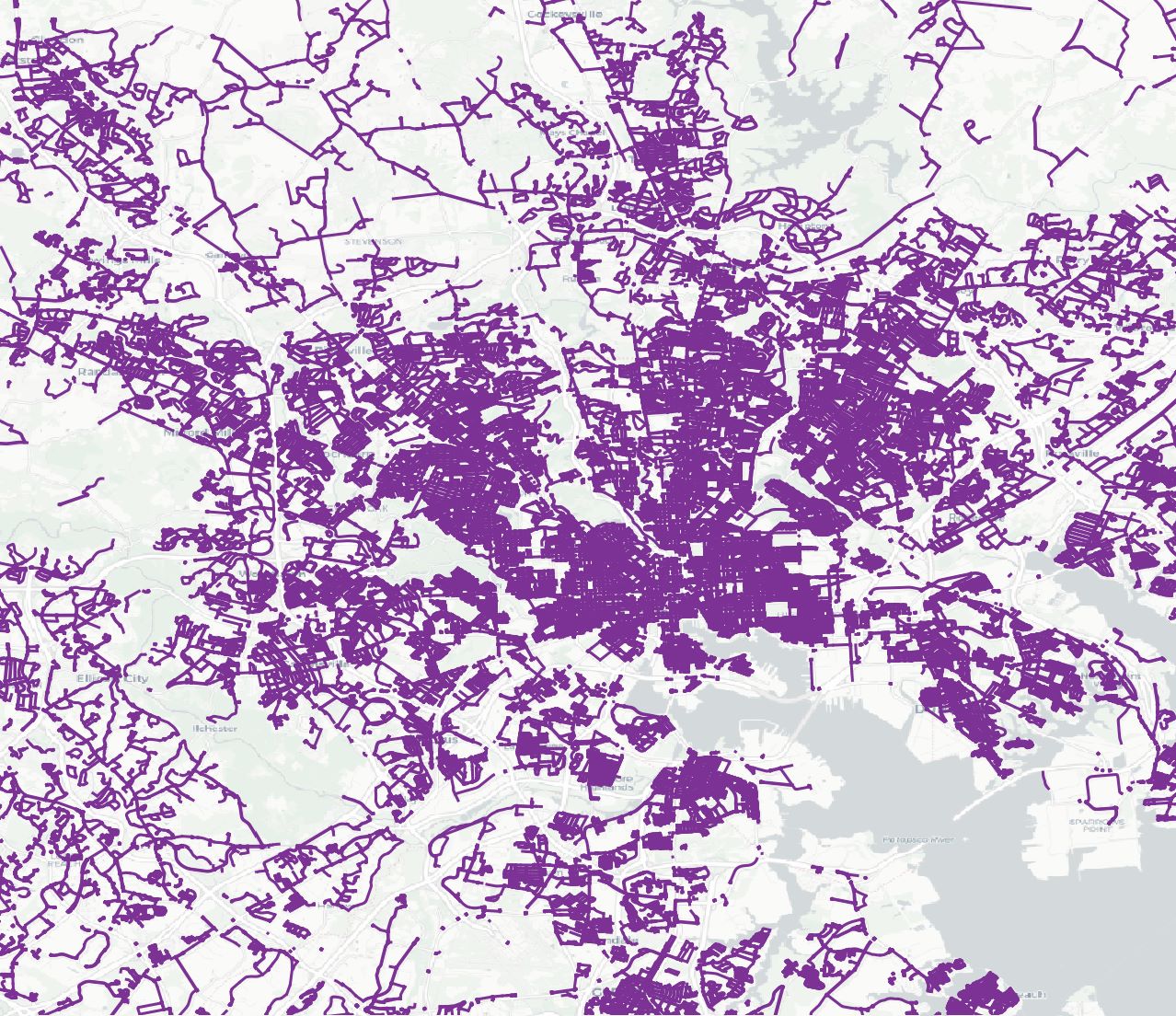}
            \caption{Baltimore County, MD}
        \end{subfigure}
    \caption{Pedestrian pathway graphs data generated with Prophet}
    \label{fig:Prophet_data_coverage}
    \end{figure}

After demonstrating the robust performance of Prophet in multiple test areas, we used Prophet to generate pedestrian pathway graphs at scale. Specifically, We generated connected pedestrian pathway graphs for 6 counties in the U.S., including (1) King County, WA (2) Snohomish County, WA (3) Multnomah County, OR (4) Columbia County, OR (5) Harford County, MD (6) Baltimore County, MD. Figure \ref{fig:Prophet_data_coverage} illustrates the prediction graphs, and Table \ref{tab:Prophet_data_stats} shows their aggregate statistics. These graphs generated by Prophet can fill the information gap in places where no other routable pathway graphs are available, and can be refined with Skeptic for more accurate representation for the actual pedestrian pathway network.

\begin{minipage}[t!]{0.85\textwidth}
    \centering
    \captionsetup{width=0.99\textwidth}
    \captionof{table}{Statistics on data generated with Prophet for U.S. counties}
    \label{tab:Prophet_data_stats}
    \resizebox{\textwidth}{!}{%
        \begin{tabular}{l|ccccc}
        \toprule
        \textbf{Area} &  \textbf{Number of sidewals} & \textbf{length of sidewalks (m)} & \textbf{Number of crossings} & \textbf{length of sidewalks (m)} & \textbf{Number of intersections} \\
        \midrule
        \textbf{King} &  249683 & 15850008 & 133603 & 1968246 & 45971\\
        \textbf{Snohomish} & 62991 &  5033712 & 33483 & 480327 & 12388\\
        \textbf{Multnomah} & 178631 &  10647695 & 92788 & 1385683 & 31697\\
        \textbf{Columbia} & 17762 &  2253858 & 9291 & 139957 & 3231\\
        \textbf{Harford} & 60979 &  5141570 & 32224 & 460859 & 11110\\
        \textbf{Baltimore} & 159842 &  11379123 & 84479 & 1215692 & 28432\\
        \bottomrule
        \end{tabular}
    }
    \vspace{2mm}
\end{minipage}\hfill

\section{Conclusions and Future Work}
\label{sec:dis}
In this work, we introduce a sociotechnical protocol combining automated inference with manual review and community engagement to achieve routable, reliable, and reproducible pedestrian pathway graphs. We piloted this protocol in WA State, demonstrating that Prophet outperforms imagery-only state-of-the-art methods, but that manual review with Skeptic offers further improvements.  We described a community engagement protocol to further enrich and validate the data for the needs of those most affected by quality issues.

In ongoing work, we are implementing feedback loops into the process to allow vetted data to be used to continuously train and improve the model and design redundant review to verify inter-mapper consistency.

\section{Acknowledgements}
This work was funded in part by the Taskar Center for Accessible Technology, USDOT ITS4US NOFO No: 693JJ322NF00001, and Microsoft's AI4Accessibility award. 

\section{Author Contribution}
The authors confirm contribution to the paper as follows: study conception and design: Yuxiang Zhang, Bill Howe, Anat Caspi; data collection: Yuxiang Zhang,  Bill Howe, Anat Caspi; analysis and interpretation of results: Yuxiang Zhang,  Bill Howe, Anat Caspi; draft manuscript preparation: Yuxiang Zhang,  Bill Howe, Anat Caspi. All authors reviewed the results and approved the final version of the manuscript.

\newpage

\bibliographystyle{trb}
\bibliography{trb_template}
\end{document}